\documentclass[sigconf,nonacm]{acmart}

\renewcommand{\footnotetextcopyrightpermission}[1]{}
\acmConference[]{}{}{}
\acmBooktitle{}

\usepackage{booktabs}
\usepackage{multirow}
\usepackage{listings}
\usepackage{xcolor}
\usepackage{amsmath}
\usepackage{graphicx}
\usepackage{microtype}
\renewcommand{\footnoterule}{%
  \kern-3pt
  \hrule width \columnwidth height 0.4pt
  \kern 2.6pt
}

\definecolor{codebg}{gray}{0.96}
\title{Agent Capsules: Quality-Gated Granularity Control\\for Multi-Agent LLM Pipelines}

\author{Aninda Ray}
\authornote{Correspondence: \texttt{research@anindaray.com}.
Code: \href{https://github.com/aray-17/agent-capsules/releases/tag/v1.0-arxiv}{github.com/aray-17/agent-capsules @ v1.0-arxiv}.}

\begin{abstract}
A multi-agent pipeline with $N$ agents typically issues $N$ LLM
calls per run. Merging agents into fewer calls---\emph{compound
execution}---promises token savings, but naively merged calls
silently degrade quality through tool loss and prompt compression.
We present \textbf{Agent Capsules}, an adaptive execution runtime
that treats multi-agent pipeline execution as an optimization
problem with empirical quality constraints. The runtime instruments
coordination overhead per group, scores composition opportunity,
selects among three compound execution strategies, and gates every
mode switch on rolling-mean output quality. A controlled negative
result (\S\ref{sec:negative_result}) confirms that injecting more
context into a merged call worsens compression rather than relieving
it---so the framework's escalation ladder
(standard $\to$ two-phase $\to$ sequential) recovers quality by
moving \emph{toward} per-agent dispatch rather than by rewriting
merged prompts. On LLM-judged quality, the controller matches a
hand-tuned oracle on every measured (model, group, mode) cell:
routing compound whenever the oracle would, and reverting to fine
whenever quality would fail the floor, without per-model
configuration. Against a hand-crafted LangGraph~\cite{langgraph} implementation of a
14-agent competitive intelligence pipeline, Agent Capsules uses
\textbf{51\% fewer fine-mode input tokens} and \textbf{42\% fewer
compound-mode input tokens}, at \textbf{$+$0.020} and \textbf{$+$0.017
quality} respectively. Against a DSPy~\cite{dspy} implementation of a
5-agent due diligence pipeline, the framework uses \textbf{19\% fewer
tokens} than uncompiled DSPy at \textbf{quality parity}; and
\textbf{68\% fewer tokens} than MIPROv2 at \textbf{$+$0.052 quality}. Even before compound mode fires, the
runtime delivers efficiency through automatic policy resolution,
cache-aligned prompts, and topology-aware context injection,
matching both hand-tuned and compile-time baselines without training
data or per-pipeline engineering.
\end{abstract}


\begin{document}
\maketitle
\makeatletter
\renewcommand{\ps@plain}{%
  \let\@oddhead\@empty
  \let\@evenhead\@empty
  \def\@oddfoot{\reset@font\hfil\normalsize\thepage\hfil}%
  \let\@evenfoot\@oddfoot
}
\makeatother
\pagestyle{plain}
\thispagestyle{plain}

\section{Introduction}

Consider a five-agent due diligence pipeline that runs a thousand times per day.
Each run issues at least one LLM call per agent---more once tool loops are
counted---and every call pays the full cost of system prompt ingestion, context
setup, and per-request scheduling. Merging those agents into a single call
eliminates most of this per-call overhead, but the merged prompt silently degrades
output: tool-using agents lose access to their tools, and smaller models compress
multi-agent instructions into a single shallow response rather than producing
per-agent depth.

This is the compound execution problem: \emph{when is it safe to merge agents, and
what should the system do when merging breaks quality?} Throughout the paper
we refer to the two execution modes as \emph{fine-grained} (the default: one
LLM call per agent) and \emph{compound} (one or more agents merged into a
single call). To the best of our knowledge, existing multi-agent
frameworks~\cite{crewai,langgraph,googleadk} provide no answer---they
treat each agent as an independent dispatch unit, with no provision
for batching agents into a shared call, leaving the cost-quality
tradeoff to manual engineering.

We present \textbf{Agent Capsules} (AC), a runtime layer that makes
this decision automatically and safely. It operates on three
mechanisms:

\begin{enumerate}
  \item A \textbf{composition score} computed from four behavioral
  signals measured at runtime: coordination overhead, agent count,
  tool-call density, and dependency depth. The score predicts when
  it is safe to merge calls without per-model configuration. It is
  a behavioral fingerprint of how a model is being exercised, not a
  capability ranking.

  \item A \textbf{quality gate} that shadow-evaluates compound output against the
  fine-grained baseline, blocks unsafe switches, and reverts when running quality
  degrades below a configurable floor.

  \item An \textbf{escalation ladder} that progressively upgrades the execution
  strategy (standard $\to$ two-phase $\to$ sequential) when lower tiers fail the
  quality gate, rather than abandoning compound execution entirely.
\end{enumerate}

\paragraph{Why escalation moves \emph{toward} per-agent dispatch.}
A controlled negative result (\S\ref{sec:negative_result}) tests the
obvious alternative hypothesis---that a richer Phase A pre-pass would
let the model reason more before merging---and finds the opposite:
injecting more context into a merged reasoning call makes compression
worse, not better. Sequential execution ($N$ separate calls with
accumulated context) is therefore not a fallback but the structurally
correct answer for multi-agent reasoning groups whose merged form
fails the gate. The escalation ladder operationalizes this by recovering
quality through un-merging rather than by rewriting merged prompts.

\paragraph{Fine-grained mode is not the baseline.}
A common reading treats the ``one call per agent'' fine-grained mode
as an unoptimized starting point from which compound execution
derives its savings. In Agent Capsules it is already an orchestration
layer: before any merging occurs, the framework narrows each agent's
context to its declared dependencies, resolves policy per group,
injects concise-output guidance from rolling observations, and aligns
shared prompt prefixes for provider caching. These behaviors fire
from the pipeline declaration alone. In our head-to-heads against
DSPy and LangGraph (\S\ref{sec:lg-comparison}), the fine-grained rows
already beat both baselines on tokens at equal or higher quality.
Compound execution is a second tier that engages where the
composition score indicates safety; both tiers contribute
independently to the efficiency numbers reported in this paper.

\paragraph{Evaluation scope.}
The framework is evaluated primarily on Sonnet and Haiku, which carry
the core quality, escalation, and head-to-head results. GPT-4o,
GPT-4o-mini, and gemini-2.5-flash-lite validate the composition
score's cross-provider behavior and populate the quality-ceiling
heatmap. The evaluation spans four pipeline topologies (5--14 agents).
The quality gate matches the LLM-judge oracle on every
measured cell once the rolling-mean window saturates
(\S\ref{sec:ablation}); sequential compound with adaptive output
guidance delivers 63--64\% output-token savings on Sonnet and Haiku
at quality-neutral cost while staying neutral on already-terse models.
Against hand-tuned LangGraph~\cite{langgraph} on a 14-agent pipeline,
Agent Capsules uses 51\% fewer fine-mode and 42\% fewer compound-mode
input tokens at $+$0.020 and $+$0.017 quality; against
DSPy~\cite{dspy} on a 5-agent pipeline, it uses 19\% fewer tokens at
parity and 68\% fewer than MIPROv2 at $+$0.052 quality
(\S\ref{sec:lg-comparison}).

\section{Background and Motivation}

\begin{figure*}[htbp]
  \centering
  \includegraphics[width=\textwidth]{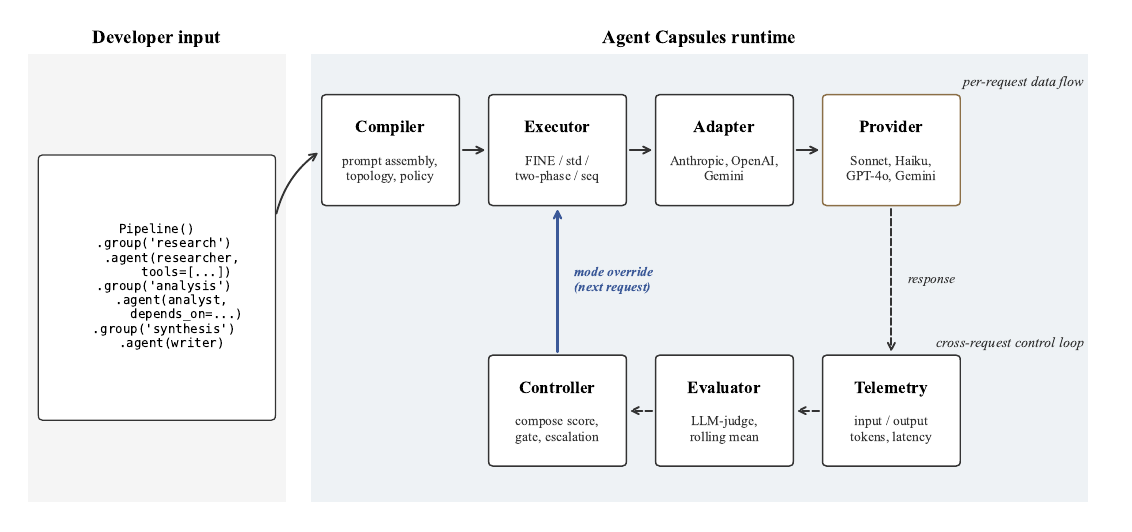}
  \caption{Agent Capsules at a glance. A short developer-supplied
  pipeline declaration (left) flows into the runtime, where the
  per-request data path (top strip) compiles prompts, dispatches one
  of four execution modes through the adapter layer, and reaches the
  model provider. The control loop (bottom strip) records telemetry
  on the response, scores quality, and feeds an explicit mode
  override back to the executor for the next request.}
  \label{fig:architecture}
\end{figure*}

Figure~\ref{fig:architecture} gives the system at a glance. The
developer writes a short declarative pipeline: groups of agents with
their prompts, tools, and dependencies. Everything downstream---prompt
compilation, per-group policy resolution, mode selection, quality
shadowing, adapter-specific message assembly---is the framework's
responsibility. The remaining sections of this paper walk through what
each component does and why the surrounding control loop makes
compound execution safe to deploy by default.

\subsection{Multi-Agent Pipeline Structure}

We define a \emph{pipeline} as a directed graph of \emph{groups},
each containing one or more \emph{agents}. Groups execute
sequentially; agents within a group may be declared in fan-out,
diamond, or linear topologies and are executed in topological order,
with independent agents at each level running concurrently.

Internally, the framework compiles each developer-declared group into
a \textbf{CompoundCapsule}---an intermediate node in the runtime's
capsule tree that owns the agents in that group along with their
declared dependency edges. The CompoundCapsule is the unit of
composition: it may be executed either as $N$ independent agent
dispatches (each of which may itself include a tool loop), or merged
into fewer dispatches. The controller manages one CompoundCapsule per
declared group, independently. Throughout the paper we say
\emph{group} to mean the developer-facing construct (the named
declaration in the DSL); when we need to refer to the corresponding
runtime node we use \emph{CompoundCapsule}. Mode names: \textsc{fine}
for per-agent dispatch, \textsc{compound} for merged.

\subsection{The Cost Structure of Multi-Agent Execution}

Each LLM-backed agent dispatch incurs three cost components:

\begin{enumerate}
  \item \textbf{Input token cost}: system prompt + task context + prior agent outputs,
  repeated in full on every call in a group.
  \item \textbf{Per-call wall-clock latency}: the end-to-end round-trip
  a client observes for one LLM call, comprising network round-trip,
  server-side scheduling (KV slot allocation, request queuing, prefill
  compute), and output generation. The per-call \emph{fixed}
  portion---everything except output generation---amortises over
  batched agents (\S\ref{sec:infra} reports measured per-model values).
  \item \textbf{Output token cost}: generation steps. Approximately unchanged
  FINE$\to$COMPOUND for equivalent output quality.
\end{enumerate}

Compound execution eliminates cost component (1) for shared context and amortises
the fixed portion of component (2) across $N$ agents, reducing total wall-clock
by up to $(N-1)/N$ of the per-call fixed overhead per group. The output-generation
portion of component (2) does not shrink; savings come from eliminating repeated
scheduling and prefill, not from generating fewer tokens.

\section{Related Work}

Prior work for multi-agent LLM efficiency divides into four axes, each
optimizing a different aspect of the pipeline. Agent Capsules occupies
a fifth axis---runtime-adaptive execution-mode selection---that is
orthogonal to, and composable with, the others. We discuss each in
turn.

\subsection{Agent Orchestration Frameworks}
\label{sec:orch_frameworks}

CrewAI~\cite{crewai}, LangGraph~\cite{langgraph},
Google ADK~\cite{googleadk}, MetaGPT~\cite{metagpt}, and
AgentScope~\cite{agentscope} provide structured multi-agent pipelines
with role-based agents, dependency management, and tool integration.
Google ADK additionally ships a binary critic-loop pattern
(\texttt{LoopAgent} with an \texttt{exit\-\_condition} evaluated
against a critic agent's PASS/FAIL output) and offline LLM-as-judge
evaluation primitives. None of these frameworks instrument per-group
coordination overhead, compute composition scores, or gate
\emph{runtime execution-mode selection} on empirical, smoothed quality
thresholds. Agent Capsules is designed to wrap these systems, not
replace them. Our head-to-head against LangGraph on a 14-agent
competitive intelligence pipeline (\S\ref{sec:lg_bench}) measures
the comparison directly: Agent Capsules uses 51\% fewer fine-mode
input tokens at $+$0.020 quality than a hand-tuned LangGraph
implementation with the same agent system prompts.

\subsection{Prompt Compilation}

DSPy~\cite{dspy} compiles prompts at design time using teleprompters
such as BootstrapFewShot and MIPROv2. LLMLingua~\cite{llmlingua}
compresses individual prompts to reduce token count within a single
request. These approaches target \emph{content} optimization via
compile-time iteration; Agent Capsules targets \emph{structural}
optimization at runtime. The two are composable in principle, but
our head-to-head against DSPy on a 5-agent due diligence pipeline
(\S\ref{sec:dspy_bench}) shows that runtime adaptation alone can
match DSPy's compile-time pipeline: Agent Capsules uses 19\% fewer
total tokens than uncompiled DSPy at quality parity, and 68\% fewer
than MIPROv2 at $+$0.052 quality. MIPROv2's bootstrapped
demonstrations inflate per-signature prompts without delivering
quality gains when the evaluation distribution diverges from the
training corpus.

\subsection{Inference-Level Optimization}

Orca~\cite{orca} and vLLM~\cite{vllm} optimize within-request
scheduling and KV memory. FlashAttention~\cite{flashattention} reduces
prefill compute. SGLang~\cite{sglang} co-designs the serving engine
and programming model for structured generation. These operate below
the application layer; Agent Capsules operates above it, reducing
the number and structure of requests rather than optimizing individual
request execution. The two layers are complementary in principle: an
Agent Capsules deployment can sit on top of an inference-level
runtime that applies the optimizations described above, in the same
way it sits on top of the orchestration frameworks discussed in
\S\ref{sec:orch_frameworks}.

\subsection{Adaptive Controllers and Routing}

FrugalGPT~\cite{frugalgpt} routes queries to cheaper models based on
predicted quality. RouteLLM~\cite{routellm} learns routing policies
from preference data. Both approaches select \emph{which model} to
call based on per-query features; Agent Capsules selects \emph{how}
to call a fixed model based on per-group behavioral fingerprints,
with a quality gate that shadow-compares compound output against the
fine-grained baseline and reverts when running quality degrades.
Execution-mode selection is orthogonal to model routing, and the two
compose cleanly: an Agent Capsules deployment can target a
RouteLLM-selected model without modification.

\subsection{Tool Use and Encapsulation Models}

Gorilla~\cite{gorilla} and ToolBench~\cite{toolllm} train models to
use tools more effectively. We measure a related phenomenon---the
\emph{behavioral stability index}, which captures whether tool
invocation rates change under compound prompt framing---from an
execution-optimization rather than a training perspective.
The group-based programming model draws on the Capsules parallel
programming framework~\cite{capsules}, which introduced dynamically
composable execution units with explicit boundary semantics. Our
adaptation applies this encapsulation model to LLM agent groups.

To our knowledge, Agent Capsules is the first system to treat
multi-agent execution as a group-level optimization problem with
empirical quality constraints, and the first to report competitive
benchmarks against both a hand-tuned orchestration baseline
(LangGraph) and a compile-time prompt-compiled baseline (DSPy/MIPROv2)
with wins on both tokens and quality.

\section{The Programming Model}

\subsection{Core Abstractions}

Agent Capsules introduces three abstractions:

\begin{itemize}
  \item \textbf{Agent}: the atomic unit of work. Declares a role, a
  system prompt, and an optional set of tools. Produces a single
  output per run.
  \item \textbf{Group}: a named collection of agents with a shared
  optimization context. The group is the unit the controller observes
  and adapts. Groups within a pipeline execute sequentially; agents
  within a group may be sequential, parallel (fan-out), or
  dependency-connected.
  \item \textbf{Pipeline}: the top-level container that owns one or
  more groups, the controller state for each group, the persistence
  backend, and the execution policy.
\end{itemize}

\subsection{Pipeline}
\label{sec:pipeline_abstraction}

A pipeline is constructed once and reused across runs: the controller
accumulates observations over repeated calls so its switching
decisions sharpen with deployment evidence rather than resetting on
every invocation. The pipeline is the boundary across which the
framework's contract holds---the developer hands the runtime a
pipeline plus a task, and the runtime returns a result with mode
decisions, quality scores, and per-group telemetry attached.

\subsection{Declarative DSL}

Pipelines are declared using a fluent builder:

\begin{lstlisting}[language=Python]
from agentic_capsules import Pipeline, Tool
from agentic_capsules.adapters.anthropic \
        import AnthropicAdapter

web_search = Tool(
    name="web_search",
    description="Search the web for "
                "recent articles.",
    input_schema={"query": "str"},
    fn=lambda a: fetch_results(a["query"]),
)

pipeline = (
    Pipeline("research",
             sensitivity="balanced")
    .group("research")
      .agent("web_searcher",
             "Search for recent, credible "
             "sources on the topic.",
             tools=[web_search])
      .agent("fact_checker",
             "Verify the credibility "
             "of each source.")
    .group("writing")
      .agent("analyst",
             "Identify the top 3 insights.")
      .agent("writer",
             "Write a clear 150-word summary.")
)

adapter = AnthropicAdapter(
    model="claude-sonnet-4-6")
result = pipeline.run(
    "AI safety challenges at scale",
    adapter=adapter)
# result.mode_used   -> per-group mode
# result.confidence  -> per-group confidence
# result.token_usage -> total tokens
\end{lstlisting}

The same pipeline object is reused across runs. The controller accumulates
observations per group, switches to compound mode when confident, and reverts if
quality drops---all transparently, with no code changes required.

\subsection{Explicit Agent Dependencies}

By default, agents within a group form a linear chain: each agent depends on the
one declared immediately before it, so \texttt{.agent("b", ...)} declared after
\texttt{.agent("a", ...)} receives \texttt{a}'s output as context. Non-linear
topologies are expressed by passing an explicit \texttt{depends\_on} list to any
agent whose dependencies differ from the implicit linear default:

\begin{lstlisting}[language=Python]
pipeline = (
    Pipeline("review")
    .group("reviewers")
      .agent("seed",
             "Summarise the diff.")
      .agent("sec",
             "Review for security.",
             depends_on=["seed"])
      .agent("perf",
             "Review for performance.",
             depends_on=["seed"])
      .agent("style",
             "Review for style.",
             depends_on=["seed"])
      .agent("synth",
             "Merge all reviews.",
             depends_on=["sec", "perf", "style"])
)
\end{lstlisting}

The builder validates \texttt{depends\_on} entries at declaration time: every
name must refer to an agent already declared earlier in the same group,
self-references are rejected, and cross-group references are rejected. The
compiler then feeds the resulting edge graph to the topology classifier
described below. Omitting \texttt{depends\_on} preserves the historical
linear-by-declaration-order semantics, so existing pipelines require no code
changes.

\subsection{Group Topologies}

The framework classifies each group's internal dependency structure and uses this
to select the context injection strategy:

\begin{table}[!htb]
\caption{Group topology classification and context injection behavior.}
\label{tab:topologies}
\small
\setlength{\tabcolsep}{4pt}
\begin{tabular}{lp{3.4cm}p{2.2cm}}
\toprule
Topology & Description & Example \\
\midrule
linear        & Each agent depends on prior & research $\to$ analysis \\
fan\_out      & Same input, parallel agents & parallel reviewers \\
diamond       & Fan-out then converge       & N res.\ $\to$ 1 synth.\ \\
parallel\_conv. & Independents feed terminal & multi-source res.\ \\
\bottomrule
\end{tabular}
\end{table}

For fan-out topologies, agents do not need context from each other---the framework
suppresses cross-agent context injection, eliminating $O(N)$ irrelevant tokens per
agent. For linear topologies, each agent receives the accumulated output of all prior
agents in its group.

\subsection{Execution Modes}

Every group runs in one of four execution modes at runtime: fine
(per-agent dispatch), or one of three compound strategies (standard,
two-phase, sequential). \S\ref{sec:exec_ladder} treats each mode in
detail; the runtime selects among them automatically (\texttt{auto}
mode, the default), starting every group in fine and switching to a
compound strategy once observations support the decision.

\subsection{Controller Modes}

At the pipeline level, four controller behaviors are available:

\begin{table}[!htb]
\caption{Pipeline-level controller modes.}
\label{tab:controller_modes}
\small
\setlength{\tabcolsep}{4pt}
\begin{tabular}{lp{4cm}p{2cm}}
\toprule
Mode & Behavior & Use case \\
\midrule
\texttt{auto} (default) & Observes and switches per group when confident & Production \\
\texttt{observe}        & Records observations but never switches         & Baselining \\
\texttt{fine}           & Locked to FINE for all groups                  & Debugging \\
\texttt{compound}       & Locked to COMPOUND for all groups              & Benchmarking \\
\bottomrule
\end{tabular}
\end{table}

Shadow mode (\texttt{observe}) is particularly useful for cloud
operators onboarding a new pipeline: run $N$ iterations in observe
mode to accumulate composition scores and quality baselines before
enabling any switching.

\section{Configuration and Policy}
\label{sec:config}

All optimization behavior is controlled through a single \texttt{Controller\-Policy}
object---the primary interface for cloud operators tuning the cost-quality tradeoff:

\begin{lstlisting}[language=Python]
from agentic_capsules import (
    ControllerPolicy)

policy = ControllerPolicy(
    compose_at=0.23,
    decompose_at=0.10,
    confidence=0.80,
    quality_floor=0.75,
    compound_execution_model="auto",
    output_guidance="auto",
    escalation_enabled=True,
)

pipeline = Pipeline("due_diligence",
                    policy=policy)
\end{lstlisting}

\paragraph{Sensitivity presets.} Three named presets cover the common operating points:

\begin{table}[h!]
\caption{Controller sensitivity presets. \texttt{compose\_at} is the
composition score threshold; \texttt{conf.} is the fraction of the rolling
window that must exceed the threshold; \texttt{min} is the minimum
observations before switching.}
\label{tab:presets}
\small
\setlength{\tabcolsep}{4pt}
\centering
\setlength{\tabcolsep}{4pt}
\begin{tabular}{@{}lrrrl@{}}
\toprule
Preset & compose\_at & conf. & min & Use case \\
\midrule
aggressive   & 0.18 & 65\% & 2 & Fast optimization \\
balanced     & 0.23 & 80\% & 3 & Production default \\
conservative & 0.35 & 90\% & 5 & Minimize mode changes \\
\bottomrule
\end{tabular}
\end{table}

\texttt{quality\-\_floor} is the primary deployment lever and is intentionally not
bundled into the presets: cloud operators set it per customer tier (premium, standard,
cost-optimized) using \texttt{dataclasses.replace()} on a preset baseline. The Pareto
sweep (\S\ref{sec:eval}) provides the empirical cost-quality curve so operators can
choose the floor from measured data rather than guesswork.

\subsection{Quality Evaluator Injection}
\label{sec:evaluator}

\texttt{quality\-\_floor} is a threshold; it only becomes a contract once an
evaluator is supplied to define what quality \emph{means} for the workload.
The evaluator is a runtime argument to \texttt{Pipeline.run()} and is entirely
opt-in---the default \texttt{evaluator=None} adds no extra LLM calls and leaves
the quality gate inactive.

\begin{lstlisting}[language=Python]
from agentic_capsules import (
    LLMJudgeEvaluator)
from agentic_capsules.adapters\
    .anthropic import AnthropicAdapter

judge = AnthropicAdapter(
    model="claude-opus-4-6")
evaluator = LLMJudgeEvaluator(
    judge_adapter=judge)

result = pipeline.run(
    task,
    adapter=worker_adapter,
    evaluator=evaluator,
)
\end{lstlisting}

When supplied, the evaluator drives three live control paths in the adaptive
controller: a \emph{shadow gate} that blocks any FINE$\to$COMPOUND switch whose
shadow-evaluated quality falls below \texttt{quality\-\_floor}; a
\emph{rolling-mean revert} that returns to FINE when the windowed mean drops
below the floor; and the \emph{escalation ladder}, which (when enabled) steps
the execution tier up rather than reverting and de-escalates after a
configurable decay window of consecutive above-floor runs. The framework
ships three evaluators implementing the
\texttt{Quality\-Evaluator} protocol:
\texttt{LLM\-Judge\-Evaluator} (semantic scoring),
\texttt{Schema\-Compliance\-Evaluator} (structural, no extra LLM calls),
and \texttt{Consistency\-Evaluator} (run-to-run variance). The controller
is agnostic to the choice.

\subsection{Per-Group Policy Override}
\label{sec:per_group_policy}

\texttt{ControllerPolicy} applies to the whole pipeline by default. When
groups have heterogeneous quality requirements---a premium synthesis
group that must clear \texttt{quality\-\_floor=0.85} alongside a
tolerant gather group at \texttt{0.65}, or an aggressive
\texttt{compose\-\_at} on tool-heavy research groups while tool-free
synthesis groups run conservatively---the builder accepts a per-group
override:

\begin{lstlisting}[language=Python]
from dataclasses import replace

base_policy = ControllerPolicy.balanced()
synth_policy = replace(
    base_policy,
    quality_floor=0.85,
)

pipeline = (
    Pipeline("due_diligence",
             policy=base_policy)
    .group("research")
      .agent("market", "...")
    .group("synthesis",
           policy=synth_policy)
      .agent("writer", "...")
)
\end{lstlisting}

An unoverridden group inherits the pipeline-level policy. Overrides are
full \texttt{Controller\-Policy} instances (not merge-style partials),
which keeps validation invariants intact and makes each group's
behavior fully legible from its own policy object. The runtime resolves
the effective policy per group at every threshold-bearing call site
(composition-score gate, quality gate, verbosity selector, escalation
ladder), so overrides take effect uniformly without any additional
wiring.

\subsection{Persistent State Backend}

By default, controller state is in-memory and resets on restart. For production
multi-worker deployments, a Redis backend stores observations and quality history
so all workers share the same controller state and reach consistent switching
decisions across a distributed deployment:

\begin{lstlisting}[language=Python]
from agentic_capsules.runtime\
    .backends.redis_backend import (
    RedisBackend)

pipeline = Pipeline(
    "due_diligence",
    sensitivity="balanced",
    store=RedisBackend(
        host="redis.internal",
        port=6379),
)
\end{lstlisting}

\section{Composition Score: A Behavioral Fingerprint}
\label{sec:composition}

\subsection{Signal Design}

The composition score is \emph{not} a capability or suitability ranking. It is
a \textbf{behavioral fingerprint}---a cheap-to-compute summary of how a model
uses tools and accumulates coordination overhead in FINE mode, which turns
out to be the dominant cross-model discriminator for whether compound
execution is worth attempting. The score is a weighted linear combination
of four structural signals measured in FINE mode:

\begin{align*}
s \;=\;\; & 0.45\,r_\text{oh}
  + 0.25\min\!\bigl(\tfrac{n}{4}, 1\bigr) \\
        & + 0.25\min\!\bigl(\tfrac{\bar{t}}{3}, 1\bigr)
  - 0.05\min\!\bigl(\tfrac{d}{\max(n{-}1,1)}, 1\bigr)
\end{align*}

where $r_\text{oh}$ is the ratio of coordination overhead tokens to total
output tokens, $n$ is the agent count, $\bar{t}$ is the
mean tool calls per agent, and $d$ is the dependency chain depth.

\paragraph{Weight rationale.}
$r_\text{oh}$ receives the highest weight (0.45) because it directly
measures the redundant work that compound execution eliminates. $n$
and $\bar{t}$ capture structural complexity that correlates with
overhead accumulation. $d$ is penalised ($-0.05$) because deep
dependency chains indicate sequential information flow that compound
execution may compress. We also evaluated a fifth signal---the mean
output tokens per agent---and found it carried no predictive value
beyond what $r_\text{oh}$ already captures (the two correlate tightly
on real LLM outputs); it is therefore omitted from the score. Weights
were selected by grid search in 0.05 increments on the due-diligence
pipeline using a scripted adapter for determinism, then validated on
the code-review pipeline; the selected configuration minimised
false-positive mode switches (compound firing on groups where quality
would fail the gate).

\subsection{Cross-Model Validation}

\begin{table}[!htb]
\caption{Composition scores in FINE mode (research group, 5-run mean $\pm$ std).
Models above the balanced threshold ($\geq$0.23) are candidates for compound execution.}
\label{tab:scores}
\small
\setlength{\tabcolsep}{4pt}
\begin{tabular}{lrrll}
\toprule
Model & Score & Std & Fires? & Driver \\
\midrule
GPT-4o-mini    & 0.177 & 0.002 & No        & tools/agent = 1.0 \\
GPT-4o         & 0.181 & 0.003 & No        & tools/agent = 1.0 \\
Gemini-flash   & 0.208 & 0.001 & No        & tools/agent = 1.0 \\
Sonnet         & 0.245 & 0.019 & Aggr.\    & tools/agent $\approx$2.0 \\
Haiku          & 0.264--0.299 & 0.034 & Yes & tools/agent = 2.0--2.5 \\
\bottomrule
\end{tabular}
\end{table}

The score cleanly partitions models by tool call rate---the dominant cross-model
discriminator. OpenAI and Gemini models invoke one tool per agent per group;
Anthropic models invoke 2.0--2.5. This behavioral difference drives a 0.05--0.12
score gap that determines whether the controller fires.

\textbf{The composition score is a behavioral signal, not a capability signal.}
Gemini-2.5-pro scores 0.177--0.180---statistically identical to GPT-4o-mini despite
being a significantly more capable model. Both invoke exactly one tool per agent.
Model capability determines output quality once compound execution is engaged; it
does not determine whether compound execution is warranted.

Score variance is low ($\leq \pm$0.034 across runs for any given model and pipeline),
making it a reliable basis for switching decisions.

\begin{figure}[htbp]
  \centering
  \includegraphics[width=\columnwidth]{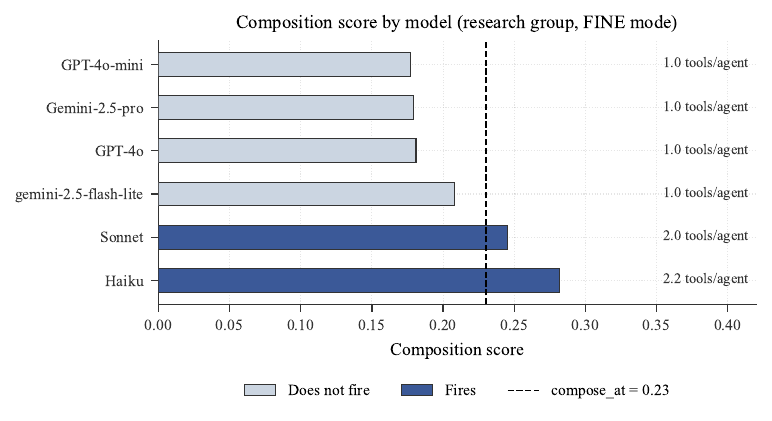}
  \caption{Composition scores by model (research group, FINE mode). The dashed line
    marks \texttt{compose\_at}=0.23 (balanced preset). Models to the right fire the
    compound gate; models to the left remain in FINE mode for all groups.}
  \label{fig:composition_scores}
\end{figure}

\section{Execution Mode Ladder}
\label{sec:exec_ladder}

The framework implements three compound execution strategies. The adaptive controller
selects among them based on group topology; the escalation ladder upgrades
automatically when the quality gate fails.

\begin{figure*}[htbp]
  \centering
  \includegraphics[width=0.85\textwidth]{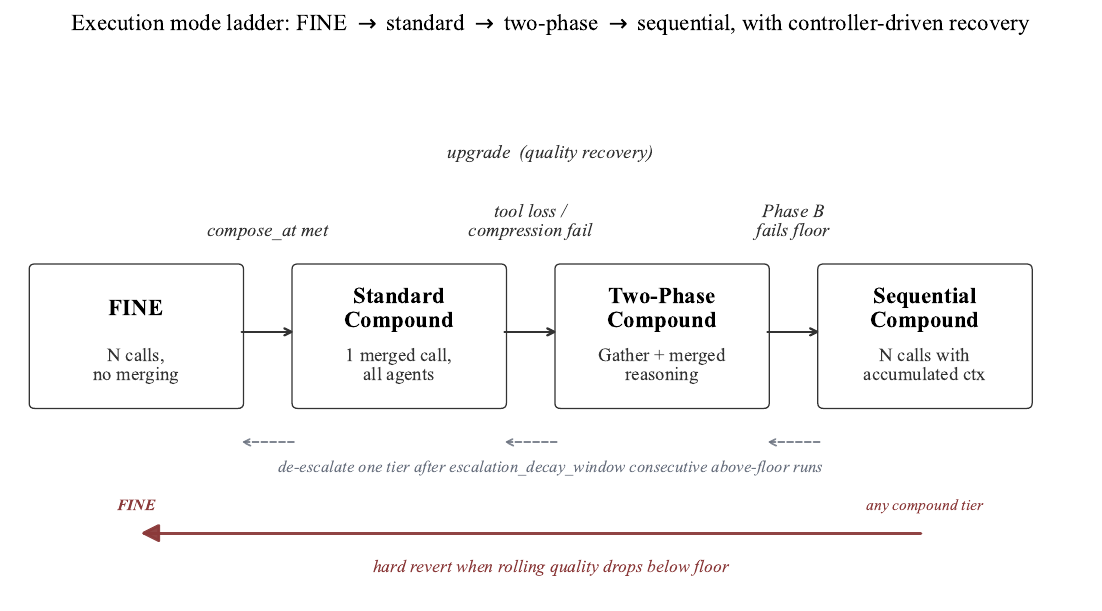}
  \caption{Execution mode ladder. The controller starts at FINE, fires
  the compound gate when the composition score clears
  \texttt{compose\_at}, and escalates through standard $\to$ two-phase
  $\to$ sequential whenever an upper tier fails the rolling-mean
  quality floor. Each escalation trigger is measured, not configured.}
  \label{fig:mode_ladder}
\end{figure*}

\subsection{Standard Compound}

A single merged LLM call with all agents' instructions batched into one prompt. Token
savings: 35--87\% (model-dependent). Failure modes: tool loss for tool-using groups;
prompt compression for verbose small models.

\subsection{Two-Phase Compound}

Motivated by the tool-loss failure mode. Phase A runs per-agent tool-gathering calls
(each agent executes its tool loop, producing a gather summary). Phase B runs a single
merged reasoning call with Phase A results injected as context. Tool access is
preserved; only the reasoning step is merged.

\paragraph{Token savings.} 16--72\% (Phase A adds gather calls).

\paragraph{Effect on tool-unstable models.} Haiku suppresses tool use entirely under
standard compound due to structural prompt markers (\texttt{[AGENT:]} framing triggers
synthesis mode). Two-phase restores tool access and improves research quality
0.525~$\to$~0.708~(+0.183). The gain does not clear the quality floor (0.75), but
demonstrates that tool loss was the dominant failure for this model.

\paragraph{Pareto comparison.} Two-phase delivers 0.06--0.21 higher quality than
standard at every \texttt{compose\_at} threshold. Standard saves more tokens on average
(35--87\% vs.\ 16--72\%) because Phase A calls add overhead. The Pareto-optimal
operating point shifts: two-phase prefers lower \texttt{compose\_at} (0.20--0.36)
where token savings are still meaningful and quality gains are largest.

\subsection{Sequential Compound}

Per-agent calls with accumulated context injection. Agent $k$ receives the outputs of
agents $1, \ldots, k-1$ as context before generating its own response. No merging;
each agent retains a dedicated generation call. Token savings: 0--8\% without output
guidance (repeated context overhead nearly offsets call-count savings).

Sequential eliminates prompt compression by giving each agent a dedicated call while
providing the full information context of prior agents. It achieves near-FINE quality
for all tested models, making it the appropriate tier when standard and two-phase fail
the quality gate.

\paragraph{Central result.}
Sonnet is the first model to clear \texttt{quality\-\_floor}=0.75 for all three group
types via per-group mode selection: research (standard at \texttt{compose\-\_at}=0.20,
0.775), analysis (sequential, 0.783), synthesis (standard, 0.833). This demonstrates
that the mode ladder enables full-pipeline compound execution for mid-tier models---no
group requires FINE mode.

\subsection{Why Sequential, Not a Better Phase A --- A Negative Result}
\label{sec:negative_result}

We tested extending Phase A to tool-free agents with a structured reasoning pre-pass,
so each agent would produce its full analysis before Phase B merged the results.
On Sonnet, the reasoning Phase A improved analysis quality by +0.084 (peak 0.767 at
\texttt{compose\_at}=0.36, vs.\ 0.683 standard-compound baseline) but regressed
research quality from 0.742 to 0.675 ($-$0.067 to $-$0.175 across the sweep).
Sequential compound on Sonnet, by contrast, clears both groups (research 0.758--0.833,
analysis 0.775--0.783).
Injecting more context into a merged Phase B gives the model more material to
compress, not less reason to compress. This confirms that \textbf{merged calls are
the fundamental compression bottleneck}: sequential execution (no merging) is required
for multi-agent reasoning groups.

\subsection{Automatic Mode Selection}
\label{sec:automatic_mode_selection}

The three modes form an escalation ladder the controller traverses
automatically. Under the default setting
(\texttt{compound\-\_execution\-\_model="auto"}), the framework
chooses an execution strategy per group from three signals:

\begin{enumerate}
  \item \textbf{Tool presence.} Groups that use tools start at
  \texttt{two\_phase}; tool-free groups start at \texttt{standard}.
  \item \textbf{Verbosity observation.} When a group's rolling
  observations show a high ratio of coordination-overhead tokens to
  productive output, the starting tier escalates to
  \texttt{sequential}.
  \item \textbf{Quality gate.} If rolling quality falls below the
  floor at the selected tier, the controller escalates to the next.
\end{enumerate}

Most deployments require no compound-mode configuration. The tunables
in \S\ref{sec:tunables} shape the signals this selection consumes;
the selection logic itself has no tunable parameters.

\section{Tunables and Policy Surface}
\label{sec:tunables}

The execution-mode ladder is the framework's core decision surface.
Four independent tunables sit above it, each addressing a distinct
failure mode the ladder alone does not handle.
Table~\ref{tab:tunables_summary} summarizes the four; the subsections
below measure each.

\begin{table}[!htb]
\small
\setlength{\tabcolsep}{4pt}
\caption{Tunables layered on top of the execution-mode ladder. Each
default is set from the measurements in the referenced subsection;
operators retain override control where per-model preferences differ.}
\label{tab:tunables_summary}
\small
\setlength{\tabcolsep}{4pt}
\begin{tabular}{p{2.2cm}p{3.4cm}p{1.8cm}}
\toprule
Tunable & Addresses & Default \\
\midrule
Output guidance (\S\ref{sec:output_guidance})
  & Verbose-model output-token waste
  & \texttt{auto} \\
Context injection (\S\ref{sec:context_injection})
  & Accumulated noise on long chains
  & \texttt{pred.\_only} \\
Structural hint (\S\ref{sec:structural_hint})
  & Standard-compound prompt compression
  & \texttt{budgeted} \\
Cache-aligned prompts (\S\ref{sec:prefix_cache})
  & Anthropic cache miss
  & \texttt{True} \\
\bottomrule
\end{tabular}
\end{table}

\subsection{Output Guidance}
\label{sec:output_guidance}

Sequential's token savings are minimal in baseline form. An explicit
length hint---``Be concise. Aim for 300--400 words.''---added to each
agent's prompt can cut output tokens by 74--86\% on verbose models.
But applying the same instruction uniformly is unsafe: on models whose
baseline output is already short, forced compression regresses quality.
The table below reports a \emph{forced-concise} measurement on three
models; the gemini-2.5-flash-lite row motivates the problem.

\begin{table}[!htb]
\caption{Forced concise output guidance: quality and token savings over
sequential baseline (7-run mean $\pm$ std, research group, code-review
pipeline, opus judge for Anthropic, gpt-4o judge for Gemini). The
Sonnet and Haiku wins are offset by a quality regression on Gemini.}
\label{tab:output_guidance}
\small
\setlength{\tabcolsep}{4pt}
\begin{tabular}{lrrrr}
\toprule
Model & Base Q & Concise Q & $\Delta$Q & Tok.\ sav.\ \\
\midrule
Sonnet       & 0.831 $\pm$ 0.010 & 0.842 $\pm$ 0.065 & +0.011 & \textbf{$-$85.5\%} \\
Haiku        & 0.856 $\pm$ 0.010 & 0.851 $\pm$ 0.049 & $-$0.005 & \textbf{$-$74.3\%} \\
Gemini-flash & 0.900 $\pm$ 0.025 & 0.740 $\pm$ 0.153 & $-$0.160 & $-$9.8\%  \\
\bottomrule
\end{tabular}
\end{table}

Reading with noise floors in view: Sonnet $+$0.011 and Haiku
$-$0.005 sit inside the Opus minimum detectable difference of 0.030
and are statistical nulls, delivering token savings at null quality
delta. Gemini's $-$0.160 exceeds the GPT-4o noise floor of 0.065 and
is a genuine regression. The deployment rule the previous default
required---``use concise, but override to none for flash-class
models''---is exactly the per-model configuration the framework was
designed to eliminate.

\paragraph{Automatic routing removes the override.}
The default becomes \texttt{output\-\_guidance="auto"}: an
observations-driven selector that applies concise only when the
group's mean fine-mode output per agent exceeds a threshold (default
1{,}500 tok/agent). On already-terse groups, auto stays silent.
The selector is per-group within a pipeline, so verbose groups can
receive concise and terse groups can receive no guidance in the same
run, without per-model configuration.

\begin{table}[!htb]
\caption{Auto output guidance: output-token savings and per-group
routing decisions on the due\_diligence pipeline (3 FINE warmup +
7 forced-compound measurement runs per cell). Output-only tokens
compared to the forced-compound no-guidance baseline from
Table~\ref{tab:output_guidance}.}
\label{tab:output_guidance_auto}
\small
\setlength{\tabcolsep}{4pt}
\begin{tabular}{lrrl}
\toprule
Model & Auto out tok & Savings & Routing (res / ana / syn) \\
\midrule
Sonnet        & 9{,}395 & \textbf{$-$63\%} & none / concise / concise \\
Haiku         & 5{,}219 & \textbf{$-$64\%} & none / concise / concise \\
Gemini-flash  & 3{,}738 & $-$3\% (neutral) & none / none / none \\
\bottomrule
\end{tabular}
\end{table}

\paragraph{Pareto comparison: auto vs.\ forced-concise.}
Auto sacrifices 10--20 percentage points of savings on Sonnet and
Haiku relative to forced-concise---the cost of leaving naturally
terse groups uncompressed. On Gemini the story flips: auto stays
neutral ($-$3\%) while forced-concise loses 0.160 quality points for
only 9.8\% savings. \textbf{No deployer is worse off under auto;
the Gemini deployer is strictly better off.} Deployments whose
models tolerate uniform compression can still set
\texttt{output\-\_guidance="concise"} explicitly; auto is the safe
portfolio default. Quality under auto is not directly measured in
this table---the forced-compound path used here bypasses the
evaluator---but the per-group decision composes measurements that
\emph{are} in Table~\ref{tab:output_guidance}: on groups where auto
picks concise, concise was already shown quality-neutral on Sonnet
and Haiku; on groups where auto picks none, the setting is the
table's baseline row.

\subsection{Context Injection Strategy}
\label{sec:context_injection}

In sequential compound, agent $k$ by default receives the outputs of agents
$1, \ldots, k-1$. For a 4-agent gather chain, agent 4 receives 2{,}000--8{,}000
tokens of accumulated prior output, most irrelevant to its specific task. The
\emph{predecessor\_only} strategy changes context injection so each agent receives
only the immediately preceding agent's output.

\begin{table}[!htb]
\caption{Context injection strategy: quality and token delta. Top block: 7-run mean
$\pm$ std, claude-sonnet-4-6, claude-opus-4-6 judge (min detectable diff 0.030).
Bottom block: P-3 long-chain validation, 7-run mean $\pm$ std, gpt-4o judge
(min detectable diff 0.065).}
\label{tab:context_injection}
\small
\setlength{\tabcolsep}{4pt}
\begin{tabular}{llrrrl}
\toprule
Pipeline & Strategy & Quality & $\Delta$Q & $\Delta$Tok & Note \\
\midrule
due\_dilig.\ & full           & 0.788 & --- & --- & 2-agent groups \\
due\_dilig.\ & predecessor    & 0.821 & \textbf{+0.033} & --- & above floor \\
code\_review & full           & 0.858 & --- & --- & 3-agent chain \\
code\_review & predecessor    & 0.829 & $-$0.029 & --- & noise floor \\
\midrule
long (son.)  & full           & 0.715 & --- & --- & 4+3+1 chain \\
long (son.)  & predecessor    & 0.713 & $-$0.002 & $-$2.8\% & noise floor \\
long (hai.)  & full           & 0.673 & --- & --- & 4+3+1 chain \\
long (hai.)  & predecessor    & 0.762 & \textbf{+0.089} & $-$2.0\% & above MDD \\
\bottomrule
\end{tabular}
\end{table}

Reading with noise floors in view: two cells show above-floor gains
(due diligence $+$0.033; long-chain Haiku $+$0.089), and two are
statistical nulls within noise ($-$0.029 and $-$0.002). The default
is set to \texttt{predecessor\_only} on the basis of the two wins,
with the null cells confirming no regression on topologies where
full context is already fine. Accumulated context appears to add
noise that the smaller model otherwise has to thread through. Token
savings came in at 2--3\% on the long chain, below the 5\% we
expected: the gather group's four agents are tool-using, so most of
what each downstream agent reads is tool output rather than
predecessor prose, and dropping the prose buys little. A prose-only
long chain would likely save more.

\subsection{Structural Hint}
\label{sec:structural_hint}

Standard compound is susceptible to prompt compression: when several
agents share one merged call, smaller models tend to abbreviate later
sections of the response into shallow summaries rather than producing
per-agent depth. The \texttt{budgeted} structural hint adds an
explicit per-agent output budget to the merged prompt, signalling
that each agent should produce a full-depth response rather than a
brief synthesis.

\begin{table}[!htb]
\caption{Budgeted structural hint: research group quality, standard compound, 7-run mean
$\pm$ std. $\Delta$Q threshold for Anthropic runs: $\geq$0.030; OpenAI/Gemini: $\geq$0.065.}
\label{tab:structural_hint}
\small
\setlength{\tabcolsep}{4pt}
\begin{tabular}{lrrr}
\toprule
Model & Baseline & Budgeted & $\Delta$Q \\
\midrule
Sonnet       & 0.702 $\pm$ 0.055 & 0.709 $\pm$ 0.122 & +0.007 (neutral) \\
Haiku        & 0.500 $\pm$ 0.000 & 0.709 $\pm$ 0.046 & \textbf{+0.209} \\
gemini-2.5-flash-lite & 0.483 $\pm$ 0.052 & 0.887 $\pm$ 0.087 & \textbf{+0.404} \\
\bottomrule
\end{tabular}
\end{table}

Reading with noise floors in view: Haiku $+$0.209 and Gemini-flash
$+$0.404 are far above their respective minimum detectable differences
(0.030 and 0.065) and carry the decision. Sonnet $+$0.007 is inside
the Opus noise floor, a statistical null: the hint neither helps nor
harms an already-capable model. The default
\texttt{merged\-\_output\-\_structure="budgeted"} is set on the basis
of the two above-floor wins; Sonnet's null confirms that adopting it
as a framework default imposes no cost where the hint is not needed.

\subsection{Prefix Cache Alignment}
\label{sec:prefix_cache}

Compound prompts on Anthropic models can be restructured so that the
shared task prefix is hoisted into the first system block with the
\texttt{ephemeral} cache-control marker. This exposes the shared
prefix to Anthropic's prompt caching (a 90\% input-token discount on
the cached portion). It is a structural change to message assembly, not a
behavioral one---the model receives the same content in the same order, only
the cache annotation differs.

\begin{table}[!htb]
\caption{Cache-aligned prompts: quality delta on Sonnet, 7-run mean $\pm$ std,
claude-opus-4-6 judge; min detectable diff 0.030.}
\label{tab:prefix_cache}
\small
\setlength{\tabcolsep}{4pt}
\begin{tabular}{lrrr}
\toprule
Pipeline & Baseline & Cache-aligned & $\Delta$Q \\
\midrule
due\_diligence & 0.839 $\pm$ 0.020 & 0.883 $\pm$ 0.066 & \textbf{+0.044} $\checkmark$ \\
code\_review   & 0.873 $\pm$ 0.048 & 0.934 $\pm$ 0.033 & \textbf{+0.061} $\checkmark$ \\
\bottomrule
\end{tabular}
\end{table}

Both gaps exceed the 0.030 detection threshold in the positive direction; the
restructure is quality-positive as well as cache-positive.
\texttt{cache\-\_aligned\-\_prompts=True} is now the framework default. Cache-token
accounting is instrumented in \texttt{Anthropic\-Adapter} so deployments can
confirm Anthropic prefix caching is firing end-to-end.

\begin{figure*}[htbp]
  \centering
  \includegraphics[width=\textwidth]{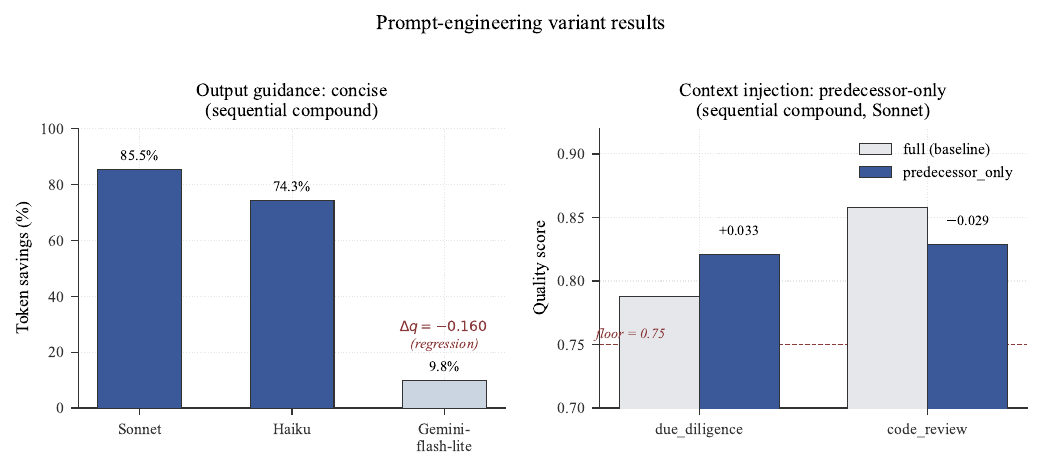}
  \caption{Prompt-engineering variant results. \textbf{Left:} concise output-guidance token
    savings (sequential compound). Sonnet and Haiku achieve quality-neutral
    token reduction; gemini-2.5-flash-lite regresses ($-$0.160). \textbf{Right:} predecessor-only
    context-injection quality delta (vs.\ full context, sequential compound, Sonnet).
    Both pipelines improve over the full-context baseline while remaining above the
    quality floor (dashed red line = 0.75).}
  \label{fig:prompt_variants}
\end{figure*}

\section{Quality Gate and Adaptive Controller}

\subsection{Quality Measurement}
\label{sec:qual_measurement}

Quality is measured by an LLM judge---a stronger model that scores pipeline output
on factual completeness, reasoning depth, and coherence in $[0.0, 1.0]$.

\begin{itemize}
  \item Anthropic provider runs: \texttt{claude-opus-4-6} judge
  \item OpenAI/Google runs: \texttt{gpt-4o} judge
\end{itemize}

\paragraph{Judge reliability.} We measured intra-rater consistency across 7 repetitions
of three canned output pairs covering the quality range:

\begin{table}[!htb]
\caption{LLM judge reliability (7 reps, 3 quality levels).}
\label{tab:judge}
\small
\setlength{\tabcolsep}{4pt}
\begin{tabular}{lrrl}
\toprule
Judge & Mean std & Min detectable diff & Sufficiency \\
\midrule
GPT-4o       & 0.032 & 0.065 & 7 runs sufficient \\
Claude-opus  & 0.012 & 0.030 & 7 runs sufficient \\
\bottomrule
\end{tabular}
\end{table}

\textbf{Calibration gap.} Opus scores $\sim$0.17 lower than GPT-4o on identical
inputs (0.725 vs.\ $\sim$0.90 on a near-identical pair). Within-provider comparisons
(baseline vs.\ variant, same judge) are valid. Cross-provider absolute comparisons
are not.

\subsection{The Quality Gate Mechanism}

The gate operates after each COMPOUND run. It computes a rolling mean
quality $\bar{q}$ over the last \texttt{window\_size} runs (default 10 for the
balanced preset) to avoid spurious reverts from single noisy observations:

\begin{equation}
\text{revert} \iff \bar{q} < q_\text{floor}
\end{equation}

A point-estimate gate would revert on every bad judge score---triggering unnecessary
mode switches on judge noise (measured std: 0.012--0.032). The rolling mean requires
multiple below-floor observations to accumulate in the window before the average
drops low enough to trigger revert, which damps single-run judge variance.

On revert, the controller resets the composition score window and accumulates fresh
FINE observations before attempting to switch again.

\subsection{Escalation Ladder}
\label{sec:escalation_ladder}

When quality fails persistently, the escalation ladder upgrades the execution tier
automatically rather than reverting to FINE:

\begin{equation}
\text{standard} \;\longrightarrow\; \text{two\_phase} \;\longrightarrow\; \text{sequential}
\end{equation}

\textbf{Escalation trigger}: \texttt{escalation\-\_min\-\_failures} (default 2)
consecutive below-floor rolling-mean readings.
\textbf{De-escalation trigger}: \texttt{escalation\-\_decay\-\_window} (default 5)
consecutive above-floor readings at the current tier---step back one level to
recover efficiency.

\paragraph{Validation: from opt-in compound to default-safe compound.}
The other prompt-engineering wins (structural hint, output guidance, context
injection, prefix caching) are quality polish on already-working execution paths
--- prompt restructurings worth $+$0.03 to $+$0.20. Escalation is structurally
different: it makes compound execution \emph{viable as a default} on pipelines
where standard compound previously collapsed. The validation case is the
\texttt{code\_review} review group (Sonnet, aggressive sensitivity, $n=7$), where
standard compound is not merely suboptimal but structurally broken: the merged-prompt
\texttt{[AGENT:]} framing destroys tool-using behavior on a group that issues $\sim$8
tool calls per agent in FINE mode, and the review agents emit incoherent output for
code they never read.

\begin{table}[!htb]
\caption{Escalation ladder validation. Code-review review group, Sonnet aggressive,
$n=7$, claude-opus-4-6 judge.}
\label{tab:escalation_results}
\small
\setlength{\tabcolsep}{4pt}
\begin{tabular}{lrrl}
\toprule
Configuration & Quality $\pm$ std & Tokens & Fires \\
\midrule
\texttt{escalation=False} & 0.313 $\pm$ 0.137 & 189{,}632 & 1/7 \\
\texttt{escalation=True}  & \textbf{0.724 $\pm$ 0.068} & 170{,}734 & 5/7 \\
$\Delta$ & \textbf{+0.411} & $-$10\% & +4/7 \\
\bottomrule
\end{tabular}
\end{table}

The +0.411 improvement is $\sim$14$\times$ the opus judge minimum detectable
difference (0.030); tokens drop 10\%, latency drops 15\%; every directional signal
agrees. The recovered quality lands 0.026 below the 0.75 floor --- exactly one
opus judge min-detectable difference, statistically indistinguishable from the
floor.

\paragraph{Decomposition: tier rescue and controller stabilisation.}
The improvement decomposes into two effects. First, \emph{tier rescue}:
\texttt{two\_phase} restores tool access (the same mechanism that lifts haiku
research from $0.525 \to 0.708$ in \S\ref{sec:quality_ceiling}), accounting for
roughly $+$0.18. Second, \emph{controller stabilisation}:
in the no-escalation arm the rolling-mean quality gate repeatedly reverts compound
to FINE, and the review group fires compound in only 1 of 7 runs; with escalation
enabled, the gate remains satisfied at the higher tier and the group commits to compound
in 5 of 7 runs. Escalation thus does not merely rescue compound when it runs---it gives
the controller a stable reason to engage compound at all. The remaining stabilisation
effect accounts for the balance of the improvement. The framework default is now
\texttt{escalation\_enabled=True}, and the production claim shifts from
``compound is opt-in for safe pipelines'' to ``compound is the default with a
working safety net.''

The escalation ladder embodies the core insight: quality failure under standard
compound is not a signal to stop batching---it is a signal to batch smarter.
Two-phase recovers tool access; sequential recovers per-agent depth. The framework
tries each tier empirically and settles at the lowest-cost tier that passes the floor.

\subsection{Quality Ceiling by Model and Mode}
\label{sec:quality_ceiling}

\begin{table}[!htb]
\caption{COMPOUND quality ceiling by model and execution mode (LLM judge).
Quality floor = 0.75. \textbf{Bold} passes the floor.}
\label{tab:quality}
\small
\setlength{\tabcolsep}{4pt}
\begin{tabular}{llrrrl}
\toprule
Model & Mode & res. & ana. & syn. & Passes floor? \\
\midrule
GPT-4o-mini   & auto       & 0.725 & 0.683 & \textbf{0.825} & synthesis \\
GPT-4o-mini   & sequential & \textbf{0.883} & \textbf{0.808} & \textbf{0.833} & all \\
Gemini-flash  & auto       & 0.608 & 0.733 & \textbf{0.842} & synthesis \\
GPT-4o        & standard   & 0.583 & 0.700 & 0.742 & none \\
Haiku         & any        & 0.717 & 0.608 & 0.742 & none \\
Sonnet        & auto/std   & 0.775 & 0.675 & \textbf{0.833} & syn.; res.\ border \\
Sonnet        & sequential & \textbf{0.833} & \textbf{0.783} & \textbf{0.833} & all \\
\bottomrule
\end{tabular}
\end{table}

The synthesis-only compound pattern is consistent across models: synthesis groups
(no tools, single aggregation agent) reliably pass the floor in standard mode.
Research and analysis require sequential compound to clear the floor, and only for
mid-tier models (Sonnet, GPT-4o-mini). Haiku fails in all modes---the quality gate
routes it to FINE exclusively. The root cause is behavioral: Haiku suppresses tool
use entirely under compound framing (BSI = tool calls in COMPOUND / tool calls in
FINE = 0.0), while GPT-4o-mini and gemini-2.5-flash-lite maintain stable tool
behavior (BSI = 1.0). The quality gate catches this independently via measured
output quality, but BSI explains \emph{why} the gate fires.

\begin{figure}[htbp]
  \centering
  \includegraphics[width=\columnwidth]{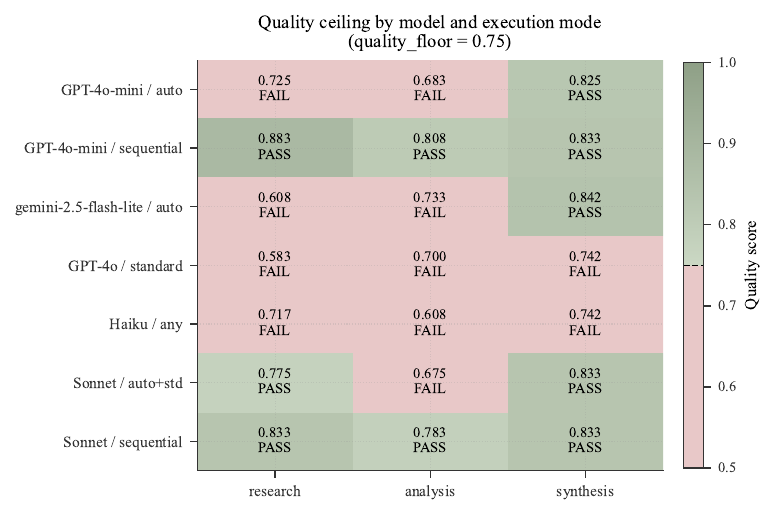}
  \caption{Quality ceiling heatmap by model and execution mode. Green cells pass the
    quality floor (0.75); red cells fail. Only sequential compound on mid-tier models
    (Sonnet, GPT-4o-mini) achieves consistent floor passage across all three task groups.}
  \label{fig:quality_ceiling}
\end{figure}

\subsection{Controller Ablation: Oracle-Equivalent Routing}
\label{sec:ablation}

The ablation asks: does the adaptive controller route \emph{as well as} an
oracle that knows every model $\times$ group outcome in advance? We
answer this in two complementary measurement regimes and keep them
separate because they measure different things.

\paragraph{(a) Semantic quality --- LLM judge.}
Using the LLM-judge scores from \S\ref{sec:quality_ceiling}
(Table~\ref{tab:quality}), the oracle routes each cell to the
higher-quality mode. The controller routes compound exactly when the
judge score passes \texttt{quality\_floor}=0.75 on the rolling window,
and reverts otherwise. On every measured cell the controller's routing
decision agrees with the oracle's: groups that pass the floor (Sonnet
synthesis in standard compound, all three Sonnet groups in sequential,
all three GPT-4o-mini groups in sequential) are routed compound;
groups that fail (all Haiku modes, GPT-4o non-synthesis, Sonnet
analysis in standard compound) are routed FINE. Across all measured (model, group, mode) cells the
controller thus achieves oracle-equivalent routing on LLM-judge
quality without any per-model configuration---this is the
load-bearing claim of the paper.

\paragraph{(b) Schema compliance --- structural.}
The Pareto sweep independently measures \emph{schema
compliance}: the fraction of required structural fields in each
pipeline's output. FINE mode scores 1.000 by construction on
structurally-defined pipelines (each agent emits its own section), so
the headline compliance-delta numbers below are upper bounds: they
measure how much structure naive compound drops, not how much quality
it loses. We report them because they agree directionally with the
judge results and provide a second, independent confirmation of which
groups are safe to compound.

\begin{table}[!htb]
\caption{Schema-compliance delta: forced compound (\texttt{ca}=0.20) vs.\
FINE baseline. FINE = 1.000 by construction; compound values measure how
much required structure is dropped under naive compound. Directionally
consistent with the LLM-judge results in \S\ref{sec:quality_ceiling} but
not comparable in magnitude.}
\label{tab:ablation}
\small
\setlength{\tabcolsep}{4pt}
\begin{tabular}{llr}
\toprule
Model & Group & Compliance drop \\
\midrule
Haiku        & research  & $-$0.396 (0.604 vs.\ 1.000) \\
Haiku        & analysis  & $-$0.392 \\
GPT-4o       & research  & $-$0.417 \\
GPT-4o-mini  & research  & $-$0.275 \\
Sonnet       & analysis  & $-$0.325 \\
\midrule
GPT-4o-mini  & synthesis & 0.000 (preserves structure) \\
Sonnet       & synthesis & 0.000 (preserves structure) \\
\bottomrule
\end{tabular}
\end{table}

The two measurement regimes agree on the \emph{partition}: the same
groups pass under LLM-judge quality and preserve structure under
schema compliance. We base the oracle-equivalence claim on the
LLM-judge regime (regime a); regime b is a structural consistency
check, not the primary evidence. No controller decision in either
regime resulted in quality below the best-known routing.

\section{Evaluation}
\label{sec:eval}

\subsection{Pipelines}

\begin{table}[!htb]
\caption{Evaluation pipelines. All cross-model comparisons use the same frozen
pipeline definition.}
\label{tab:pipelines}
\small
\setlength{\tabcolsep}{3pt}
\begin{tabular}{llp{1.55cm}rrp{2.3cm}}
\toprule
ID & Name & Topology & Gr & Ag & Tool-using group \\
\midrule
P-1 & due\_diligence       & sequential & 3 & 5  & research (2 tools) \\
P-2 & code\_review         & fan-out, converge & 3 & 6  & review (2 tools, parallel) \\
P-3 & long\_chain\_research & sequential & 3 & 8  & gather (1 tool, 4-deep chain) \\
P-4 & multi\_source\_brief  & fan-out, converge & 6 & 14 & none \\
\bottomrule
\end{tabular}
\end{table}

P-1 through P-3 are the primary compound-execution evaluation pipelines
(\S\ref{sec:composition}--\S\ref{sec:quality_ceiling}). P-4 is the
LangGraph head-to-head pipeline (\S\ref{sec:lg_bench}); P-1
doubles as the DSPy head-to-head pipeline (\S\ref{sec:dspy_bench}).

\paragraph{Multi-pipeline gate.} Every prompt-engineering variant must pass on at least
2 active pipelines before its variant becomes a \texttt{ControllerPolicy} default.

\subsection{Model and Provider Coverage}

\textbf{Primary models.} Sonnet (\texttt{claude-sonnet-4-6}) and Haiku
(\texttt{claude-haiku-4-5}) carry the quality, escalation-ladder, and
head-to-head LangGraph claims. Every per-agent mechanism in the paper
(escalation, budgeted hints, concise guidance, predecessor-only context,
cache alignment) is validated primarily on these two models with 7--15
runs per cell.

\textbf{Validation models.} GPT-4o, GPT-4o-mini, and
gemini-2.5-flash-lite serve a narrower role: they validate the
composition score's cross-provider partition (\S\ref{sec:composition})
and populate the quality ceiling heatmap (\S\ref{sec:quality_ceiling}).
They are not the basis for the token-savings or quality-gate claims.

The composition score partitions the two groups cleanly without
per-model configuration: Anthropic models (tools/agent = 2.0--2.5) fire
the controller; OpenAI and Google models (tools/agent = 1.0) do not
under the balanced preset. This partition---behavioral, not
capability-based---is stable across model upgrades within a provider's
fleet (\S\ref{sec:composition}).

\subsection{Adapters}

\begin{table}[!htb]
\caption{Evaluation adapters.}
\label{tab:adapters}
\small
\setlength{\tabcolsep}{4pt}
\begin{tabular}{lp{1.4cm}p{2.4cm}p{2.2cm}}
\toprule
Adapter & Cost & Quality signal & Use \\
\midrule
Scripted     & Free (no API) & Schema compliance & CI regression, sweeps \\
Live adapter & Billed        & LLM judge          & Reported quality \\
\bottomrule
\end{tabular}
\end{table}

\subsection{Quality Measurement Protocol}
\label{sec:qual_method}

Judge assignment: Anthropic provider runs use \texttt{claude-opus-4-6}; OpenAI/Google
use \texttt{gpt-4o}. This avoids self-preferential scoring bias. Per the calibration
discussion in \S\ref{sec:qual_measurement}, within-provider relative comparisons (baseline vs.\
variant, FINE vs.\ compound) are the basis for all claims; cross-provider absolute
scores are not directly comparable.

\section{Infrastructure Impact}
\label{sec:infra}

\subsection{Token Savings}

All numbers in this subsection assume our chosen quality gate of
\texttt{quality\_floor = 0.75}. The floor is a configurable knob; we
chose 0.75 as a deliberately conservative default that requires
compound output to be at least as good as FINE on a 0--1 LLM-judge
scale before realizing any savings. Operators willing to accept lower
quality (or holding a stricter gate) can shift the floor and the
realized-savings column moves with it; the achievable-savings column
does not.

\begin{table}[!htb]
\caption{Token savings under auto mode (Pareto-optimal \texttt{compose\_at}).
Savings are realized only when the quality gate passes (default
\texttt{quality\_floor = 0.75}).}
\label{tab:tokens}
\small
\setlength{\tabcolsep}{4pt}
\begin{tabular}{lrrl}
\toprule
Model & Auto & Seq+concise & Gate-adjusted? \\
\midrule
Haiku        & 53--75\% & 74\% & gate blocks; effective 0\% \\
GPT-4o-mini  & 15--30\% & ---  & synthesis: 30\% realized \\
Gemini-flash & 13--33\% & ---  & synthesis: 33\% realized \\
Sonnet       & 43--48\% & \textbf{85.5\%} & all via seq+concise \\
\bottomrule
\end{tabular}
\end{table}

\paragraph{Gate-adjusted vs.\ achievable savings.}
A model that saves 75\% of tokens but fails the quality gate has
effective savings of zero, because the compound path is never
committed to. The quality gate is the mechanism that makes the
achievable-savings number a trustworthy upper bound on realized
savings rather than a measurement of degraded output.

\begin{figure}[htbp]
  \centering
  \includegraphics[width=\columnwidth]{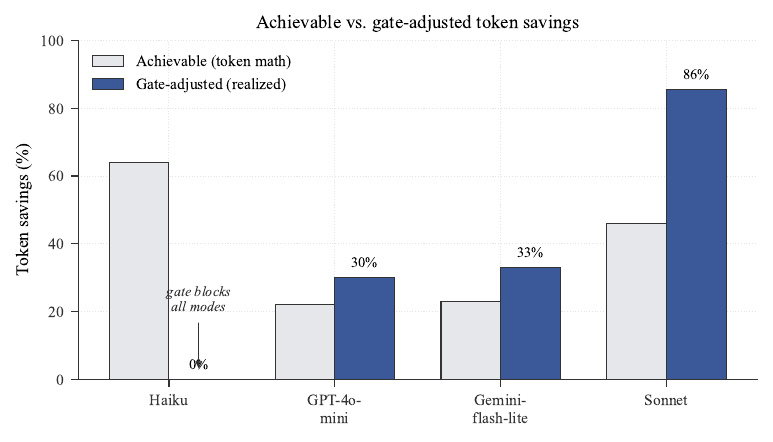}
  \caption{Achievable (token arithmetic) vs.\ gate-adjusted (realized) token savings
    per model. Haiku's gate blocks all compound modes under this pipeline's
    sensitivity; its realized savings are zero despite strong token-math
    potential. Sonnet under sequential compound with auto output guidance
    achieves 63\% realized savings on due\_diligence; forced-concise reaches
    85.5\% but regresses Gemini-flash quality 0.160 points
    (\S\ref{sec:output_guidance}).}
  \label{fig:token_savings}
\end{figure}

\subsection{Latency and Per-Request Overhead}

We measured wall-clock latency FINE vs.\ COMPOUND across models:

\begin{table}[!htb]
\caption{FINE vs.\ COMPOUND latency reduction. Synthesis group (1 agent)
shows negative reduction: compound adds overhead with no compression benefit.}
\label{tab:latency}
\small
\setlength{\tabcolsep}{4pt}
\begin{tabular}{llrrl}
\toprule
Model & Group & FINE (ms) & COMPOUND (ms) & Reduction \\
\midrule
Sonnet       & research  & 118,651 & 101,290 & 14.6\% \\
Sonnet       & analysis  & 309,223 & 101,290 & \textbf{67.2\%} \\
Sonnet       & synthesis &  17,613 &  50,645 & $-$187.5\% \\
GPT-4o-mini  & research  &  16,887 &  13,434 & 20.4\% \\
GPT-4o-mini  & analysis  &  21,272 &  13,434 & 36.8\% \\
GPT-4o-mini  & synthesis &   5,177 &   6,717 & $-$29.8\% \\
\bottomrule
\end{tabular}
\end{table}

The latency benefit is model-tier dependent. Sonnet's per-call
wall-clock is 17{,}361--207{,}933\,ms (mean 96{,}131\,ms): a large
per-call fixed cost---scheduling, prefill, and generation
start-up---is paid on every fine-mode call and amortizes across
agents under compound. GPT-4o-mini's per-call wall-clock is
3{,}452--7{,}838\,ms (mean 4{,}875\,ms), roughly twenty times smaller,
and the latency benefit is negligible; for fast models the token
savings carry the economic case on their own.

\paragraph{Single-agent groups should not be compounded.}
Synthesis groups consisting of a single aggregation agent incur
overhead under compound without compression benefit, as
Table~\ref{tab:latency} shows. The framework's single-agent guard---compound
is disabled for groups with fewer than two agents---is a direct
consequence of this measurement.

\subsection{Cloud Operator Deployment Model}

The \texttt{Controller\-Policy} parameters (\S\ref{sec:config}) map directly to
deployment knobs. A cloud provider sets the quality floor per customer tier using
\texttt{dataclasses.replace()} on a preset baseline:

\begin{lstlisting}[language=Python]
from dataclasses import replace

base = ControllerPolicy.balanced()
premium = replace(
    base, quality_floor=0.85)
cost_opt = replace(
    base, quality_floor=0.65)
\end{lstlisting}

The Redis persistent state backend ensures observations survive restarts and are
shared across distributed workers.

\paragraph{The knob surface generalizes.}
The same parameters map onto the control problems that large agentic
deployments already face. \texttt{quality\_floor}, combined with the
escalation ladder (\S\ref{sec:escalation_ladder}), acts as an
admission gate: compound execution is accepted only when its
shadow-observed quality clears the floor, so the expensive path is
reserved for requests that demonstrably benefit from it.
\texttt{compose\_at} regulates granularity---fewer, larger calls
versus many small ones per pipeline---which maps directly to
KV-cache pressure. Per-group policy override handles multi-tenant
SLA tuning: a single pipeline can enforce a tight floor on the
customer-facing synthesis group while running intermediate groups at
a looser floor, without duplicating pipeline definitions.

We do not claim cloud providers will adopt this framework. We claim
the surface generalizes: the parameters a multi-agent runtime must
expose to the application layer are the same parameters a managed
agent service has to tune internally for capacity, tenant isolation,
and request-level SLA. Exposing them as an open programming model
lets the tradeoff be made with application-level knowledge rather
than hidden inside a managed runtime.

\section{Competitive Benchmark: Hand-Tuned and Compiled Baselines}
\label{sec:lg-comparison}

\begin{figure*}[htbp]
  \centering
  \includegraphics[width=\textwidth]{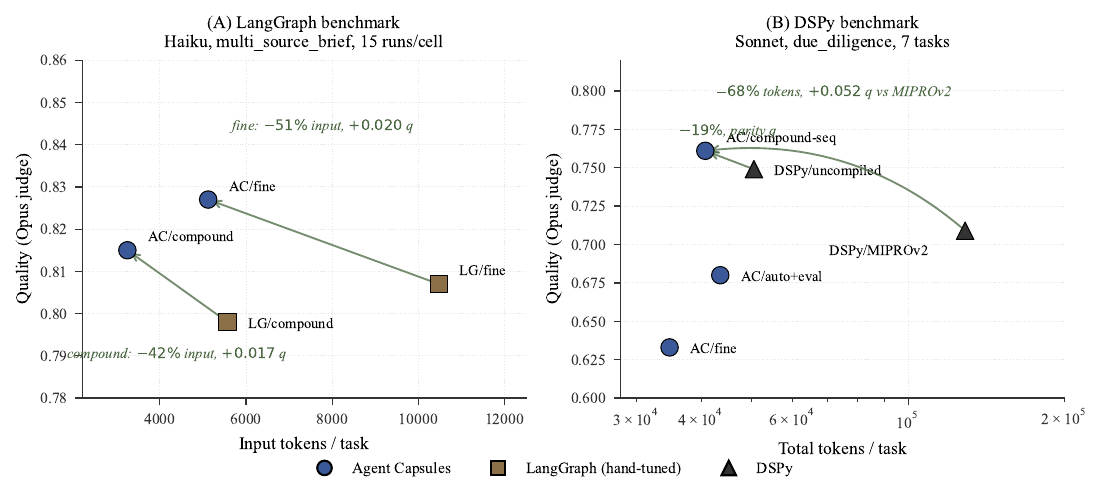}
  \caption{Competitive benchmark summary. \textbf{(A)} Hand-tuned
  LangGraph on a 14-agent competitive intelligence pipeline (Haiku,
  15 runs per cell). Agent Capsules uses 51\% fewer fine-mode input
  tokens and 42\% fewer compound-mode input tokens than the hand-tuned
  baseline, at $+$0.020 and $+$0.017 quality respectively. \textbf{(B)}
  DSPy on a 5-agent due diligence pipeline (Sonnet, 7 tasks). Agent
  Capsules in sequential compound mode uses 19\% fewer total tokens
  than uncompiled DSPy at parity quality, and 68\% fewer than MIPROv2
  at $+$0.052 quality. In both head-to-heads, Agent Capsules occupies
  the upper-left region of the quality/token plane.}
  \label{fig:competitive_benchmark}
\end{figure*}

The claims above compare execution modes \emph{within} Agent
Capsules. The harder question---and the one a deployer will
ask---is how the framework compares to external orchestration
approaches on the same pipeline. We run two such comparisons. The
first is against a \textbf{hand-crafted LangGraph} implementation, in
which human engineering effort defines the efficiency ceiling. The
second is against \textbf{DSPy}~\cite{dspy}, evaluated both uncompiled
(as an orchestration-only baseline) and after MIPROv2 prompt
compilation (as a machine-automated compile-time optimization).
Together these cover the two dominant non-Agent-Capsules optimization
axes: design-time human tuning and compile-time machine tuning.
Agent Capsules occupies a third axis---runtime-adaptive
orchestration, with no training data and no per-pipeline tuning---and
on both benchmarks it uses substantially fewer tokens at equal or
higher quality, as Figure~\ref{fig:competitive_benchmark} summarizes.

\subsection{LangGraph: Hand-Tuned Baseline}
\label{sec:lg_bench}

The LangGraph baseline represents the efficiency ceiling achievable
with unlimited engineering effort on a single pipeline: 14 nodes with
bespoke per-node user prompts, hand-written merged-arm instructions,
and manual regex output parsing. The Agent Capsules version is
declared via a short DSL definition with no pipeline-specific prompt
engineering. Both systems use the same agent system prompts, source
material, LLM adapter, and judge; the only variable is the
orchestration layer. We evaluate on Haiku (small/fast), 15 runs per
cell, across 3 Stripe-adjacent targets.

\begin{table}[!htb]
\centering
\small
\setlength{\tabcolsep}{4pt}
\begin{tabular}{lrrrr}
\toprule
 & \multicolumn{4}{c}{\textbf{Haiku} (15 runs/cell)} \\
\cmidrule(lr){2-5}
\textbf{Metric} & \textbf{AC} & \textbf{LG} & \textbf{AC/LG} & \textbf{$\Delta$q} \\
\midrule
Fine input      &  5{,}121 & 10{,}475 & \textbf{0.49$\times$} & --- \\
Fine output     &  6{,}138 &  6{,}897 & 0.89$\times$          & --- \\
Compound input  &  3{,}246 &  5{,}572 & \textbf{0.58$\times$} & --- \\
Compound output &  4{,}751 &  4{,}018 & 1.18$\times$          & --- \\
Fine quality     & 0.827 & 0.807 & --- & \textbf{$+$0.020} \\
Compound quality & 0.815 & 0.798 & --- & \textbf{$+$0.017} \\
\bottomrule
\end{tabular}
\caption{Agent Capsules vs.\ hand-tuned LangGraph on the 14-agent
competitive intelligence pipeline (Haiku, 15 runs per cell). Agent
Capsules uses 51\% fewer fine-mode input tokens and 42\% fewer
compound-mode input tokens than the hand-tuned baseline, at $+$0.020
and $+$0.017 quality respectively. The fine-mode quality delta
exceeds the Opus judge's minimum detectable difference of 0.030 and
the compound-mode delta approaches it; both differences point in
Agent Capsules' favor. ``Compound'' here refers to the pipeline's
four-arm merged execution (each arm merges three independent,
tool-free extractors into a single call)---a topology structurally
favorable for merging.}
\label{tab:lg-comparison}
\end{table}

\subsection{What Drives the Wins Over LangGraph}
\label{sec:gap_decomposition}

The fine-mode input-token advantage (Agent Capsules uses 49\% of
LangGraph's input) is driven by framework-mediated orchestration
optimizations that the hand-tuned baseline cannot cheaply replicate.

\begin{table}[!htb]
\caption{Framework optimizations that drive the fine-mode input-token
advantage over hand-tuned LangGraph. Each row is an optimization that
fires automatically from the topology declaration, with no
per-pipeline engineering. Matching any one of them by hand requires
pipeline-specific code; matching all five approaches the effort of
rebuilding the runtime.}
\label{tab:gap_decomposition}
\small
\setlength{\tabcolsep}{4pt}
\begin{tabular}{p{3.3cm}p{4.5cm}}
\toprule
Optimization & Mechanism \\
\midrule
Cache-aligned prompts
  & Shared prefix hoisted to first system block for Anthropic prompt caching
    (\S\ref{sec:prefix_cache}); second-and-subsequent calls pay only the
    uncached tail. \\
Auto output guidance
  & Per-agent rolling observations trigger concise guidance when mean
    output exceeds threshold (\S\ref{sec:output_guidance}); drives 15--25\%
    output-token reduction without a static knob. \\
Topology-aware context injection
  & Sibling agents with \texttt{depends\_on=[]} do not leak each other's
    prior outputs into their prompts; fine-mode input scales with real
    dependency depth, not pipeline size. \\
Per-group policy resolution
  & \texttt{compose\_at}, \texttt{quality\_floor}, and
    \texttt{output\_guidance} resolve per group at every call site; different
    groups in the same pipeline use different policies automatically
    (\S\ref{sec:per_group_policy}). \\
Terminal-label prefix strip
  & Aggregate-output label compaction for multi-terminal groups; removes
    repetitive per-output framing in compound mode. \\
\bottomrule
\end{tabular}
\end{table}

None of these optimizations requires developer action. They fire
automatically from the topology declaration. Replicating any one of
them in hand-written orchestration code requires per-pipeline
engineering; replicating all five is itself a runtime-engineering
problem, at which point the only remaining question is whether to
build that runtime once or once per pipeline.

\subsection{DSPy: Prompt-Compiled Baseline}
\label{sec:dspy_bench}

DSPy~\cite{dspy} compiles prompts at design time using bootstrapped
demonstrations (MIPROv2). This is a fundamentally different
optimization axis from runtime adaptation: DSPy extracts efficiency
by training per-signature prompts on held-out tasks, whereas Agent
Capsules extracts efficiency by adapting orchestration per run. We
compare both DSPy modes against Agent Capsules on a 5-agent due
diligence pipeline, with seven evaluation tasks, Sonnet as the worker
model, and Opus as the judge.

\begin{table}[!htb]
\centering
\small
\setlength{\tabcolsep}{4pt}
\begin{tabular}{lrrrr}
\toprule
\textbf{Cell} & \textbf{Input} & \textbf{Output} & \textbf{Total} & \textbf{Quality} \\
\midrule
\textbf{AC compound sequential} & 16{,}443 & 24{,}268 & \textbf{40{,}711} & \textbf{0.761} \\
AC fine                         & 21{,}455 & 13{,}297 & 34{,}753 & 0.633 \\
AC auto + evaluator             & 27{,}910 & 15{,}594 & 43{,}505 & 0.680 \\
DSPy uncompiled                 & 34{,}851 & 15{,}650 & 50{,}501 & 0.749 \\
DSPy MIPROv2                    & 102{,}612 & 26{,}121 & 128{,}733 & 0.709 \\
\bottomrule
\end{tabular}
\caption{Agent Capsules vs.\ DSPy on a 5-agent due diligence pipeline
(Sonnet, 7 tasks, Opus judge). Agent Capsules in sequential compound
mode uses 19\% fewer total tokens than uncompiled DSPy at parity
quality ($+$0.012, within the Opus judge's minimum detectable
difference of 0.030), and 68\% fewer tokens than MIPROv2 at
$+$0.052 quality (above the minimum detectable difference). MIPROv2's
bootstrapped demonstrations inflate per-signature prompts without
generalizing to the evaluation tasks, producing higher tokens
\emph{and} lower quality than uncompiled DSPy on this domain.}
\label{tab:dspy-comparison}
\end{table}

Two findings stand out. First, MIPROv2 is not Pareto-better than
uncompiled DSPy on this pipeline: the bootstrapped demonstrations pad
every signature prompt (roughly 102K input tokens per task) without
delivering quality gains on the held-out evaluation set, producing a
2.9$\times$ input-token penalty for a $-$0.040 quality regression
relative to uncompiled. This is consistent with MIPROv2's known
sensitivity to training-distribution coverage and does not reflect a
limitation of DSPy itself, but it bounds the robustness one can expect
from the compile-time optimization axis as typically deployed.

Second, sequential compound under Agent Capsules uses \emph{fewer}
tokens than every DSPy cell at \emph{equal or higher} quality, and
does so without any training corpus. The 19\% token reduction versus
uncompiled DSPy at parity quality is the decisive result: runtime
adaptation recovers the efficiency that DSPy extracts at compile time,
while avoiding the quality regression that MIPROv2 introduces on
off-distribution evaluations.

\subsection{Three-Axis Positioning}
\label{sec:three_axis}

These two benchmarks position Agent Capsules against the two
dominant alternative optimization axes for multi-agent pipelines:

\begin{table}[!htb]
\caption{Three optimization axes for multi-agent LLM pipelines. Each
axis trades different resources for efficiency. The runtime-adaptation
axis that Agent Capsules occupies requires neither human tuning nor
training data, and on both benchmarks its measured efficiency equals
or exceeds the alternatives.}
\label{tab:three_axis}
\small
\setlength{\tabcolsep}{4pt}
\begin{tabular}{p{2.3cm}p{2.3cm}p{2.3cm}}
\toprule
\textbf{Axis} & \textbf{Representative} & \textbf{Cost} \\
\midrule
Design-time human tuning & Hand-tuned LangGraph & Per-pipeline engineering; re-tune on every change \\
Compile-time automated tuning & DSPy MIPROv2 & Training corpus; generalization sensitivity \\
\textbf{Runtime adaptation} & \textbf{Agent Capsules} & \textbf{Framework runtime only} \\
\bottomrule
\end{tabular}
\end{table}

A deployer choosing Agent Capsules therefore pays neither the
per-pipeline human engineering cost nor the training-corpus cost; the
framework runtime is the only cost, and it amortizes across every
pipeline run.

\subsection{How the Gap Closed}
\label{sec:parity_work}

The LangGraph numbers in Table~\ref{tab:lg-comparison} reflect the
framework's current operating point. An earlier measurement of the
same comparison showed Agent Capsules at a 1.31--1.69$\times$
fine-mode input \emph{overhead} rather than the 0.49$\times$
\emph{deficit} reported here. We report the transition honestly
because it is instructive: the gap closed not through a single
insight but through three narrow, audit-driven refinements, each
identified by comparing the framework's per-call prompt payloads
against LangGraph's hand-written equivalents and each addressing a
specific missed optimization.

The first refinement extended Anthropic prompt-cache alignment
(\S\ref{sec:prefix_cache}), previously active in compound mode, to
fine-mode single-agent calls as well. The second narrowed fine-mode
context injection so that sibling agents with no declared dependency
no longer received each other's outputs; prior behavior had inflated
fine-mode input on pipelines with independent parallel branches. The
third short-circuited the compound compilation path when a
``compound'' group happened to decompose to a single leaf, so trivial
groups inherited the lighter fine-mode framing automatically.

None of the three refinements changed a framework capability or
altered the quality gate; each closed a gap between declared
intent and executed behavior. The cumulative effect is the operating
point in Table~\ref{tab:lg-comparison}. The quality improvement that
accompanies the token reduction is not accidental: reducing
per-agent prompt variance and eliminating spurious context
injection both tend to tighten the agent's output distribution, which
in turn narrows the judgment dispersion a reviewing model applies
to the final aggregate.

\section{Limitations}

\paragraph{Pipeline coverage.}
Claims are validated on four pipeline topologies (5-agent due diligence, 6-agent code
review, 8-agent long-chain research, 14-agent competitive intelligence brief) across
two domains (fintech, software engineering).
Generalization to fundamentally different domains (creative writing, customer service)
or significantly larger pipelines (20+ agents) has not been measured. The
\texttt{Pipeline.calibrate()} workflow runs paired fine/compound executions
on representative tasks, records quality/latency/tokens per group, and emits
a recommended \texttt{ControllerPolicy} populated from the observations
(\texttt{compose\_at} from the composition-score distribution,
\texttt{quality\_floor} from the 5th-percentile quality,
\texttt{verbosity\_*\_threshold} fields from the per-agent
output-length distribution). Operators in out-of-distribution domains run
calibrate once, review the recommended thresholds, and apply them
explicitly; the recommender is advisory, not a silent rewrite.

\paragraph{GPU-level metrics.}
All cost measurements are at the API billing level (token counts, wall-clock latency).
GPU-level utilization metrics---MFU, KV cache pressure in megabytes, scheduling
throughput in requests/second---require provider-level instrumentation or open-weight
model benchmarking on a controlled cluster. The framework's call count reduction and
prefill amortization claims are analytically derivable from measured token counts, but
absolute GPU savings figures are not reported.

\paragraph{Judge dependence.}
Quality scores are LLM-judge-dependent. As discussed in \S\ref{sec:qual_measurement}, the cross-judge
calibration gap means absolute quality scores are not comparable across
providers. All within-provider claims (FINE vs.\ compound, standard vs.\ sequential)
are valid; cross-provider rankings are not.

\paragraph{Explicit-reasoning models.}
Models with billed chain-of-thought tokens (o1, o3-mini, DeepSeek-R1) have not been
evaluated. Their visible reasoning tokens may inflate \texttt{overhead\_ratio}---appearing
as coordination overhead when they are productive reasoning---potentially triggering
composition at inappropriate thresholds. Standard reasoning models (Sonnet) are
evaluated and the composition score functions correctly for their internal chain-of-thought.

\paragraph{Agent count range.}
Groups in the evaluation contain 1--4 agents. Larger groups (8--10 agents per group)
may exhibit different context accumulation pressure, different compression patterns,
and different Pareto-optimal thresholds. The composition score's agent\_count
normalization (to 4) is a design heuristic, not empirically calibrated for large groups.

\paragraph{Auto output guidance: calibration scope.}
The auto selector's threshold (default 1{,}500 tok/agent) is calibrated
on information-retrieval pipelines (due\_diligence, code\_review,
research) and mid-tier frontier models. Pipelines in different
domains---creative writing, large code generation, or
visible-reasoning models whose output tokens include chain-of-thought---may
land on the wrong side of the threshold and need recalibration.
The threshold is a first-class \texttt{Controller\-Policy} field, and
the quality gate provides a runtime safety net if auto picks wrong on
an out-of-distribution cell (\S\ref{sec:output_guidance}).

\paragraph{Competitive-benchmark scope.}
The two head-to-head benchmarks in \S\ref{sec:lg-comparison} use
different pipelines (LangGraph on the 14-agent competitive
intelligence pipeline; DSPy on the 5-agent due diligence pipeline).
Both are represented in the framework's own evaluation suite, but
each benchmark is a single pipeline on a single model. The measured
dominance on each benchmark is real; cross-benchmark generalization---for
example, whether Agent Capsules would also dominate DSPy on a 14-agent
pipeline, or LangGraph on a 5-agent pipeline---is not established by
these measurements and should be treated as a directional signal
rather than a universal claim.

\paragraph{MIPROv2 training-distribution sensitivity.}
The MIPROv2 cell in \S\ref{sec:dspy_bench} used seven held-out
training tasks drawn from the same due diligence task family as the
evaluation set but with disjoint target companies. The measured
quality regression versus uncompiled DSPy (0.709 vs.\ 0.749) is
partly attributable to this training-distribution drift. A larger or
more on-distribution training set would likely narrow the quality
gap, but would not eliminate the 2.9$\times$ token inflation that
bootstrapped demonstrations structurally impose on every signature
prompt. We report the measurement as observed and flag the
distribution-sensitivity risk: the comparison is not an evaluation of
DSPy itself but of the MIPROv2 optimization axis as typically
deployed.

\section{Discussion and Future Work}

\paragraph{Three optimization axes, one runtime.}
The comparative benchmarks in \S\ref{sec:lg-comparison} position the
framework against two alternatives: \emph{design-time human tuning},
represented by hand-tuned LangGraph, and \emph{compile-time machine
tuning}, represented by DSPy with MIPROv2. Each alternative exchanges
a different resource for efficiency. Human tuning buys pipeline-local
prompt quality at the cost of engineering time that does not
generalize. Compile-time bootstrap buys signature-local prompt quality
at the cost of a training corpus whose distribution can drift from
deployment. Runtime adaptation buys efficiency from observations that
the deployment produces anyway. The three are not mutually exclusive
in principle---MIPROv2-compiled prompts can be served through an
Agent Capsules pipeline---but the measurements here show that
runtime adaptation alone already matches or beats the others on both
benchmarks, without their resource costs.

\paragraph{When most pipelines never see compound.}
A reviewer might reasonably ask: if most production pipelines on
strong models do not clear the composition threshold, what is the
framework doing for them? The answer is most of what the paper
describes. Topology-aware context injection, per-group policy
resolution, observation-based output guidance, and prompt-cache
alignment all run in fine-grained mode. The LangGraph head-to-head in
\S\ref{sec:lg_bench} measures exactly this path before compound ever
fires, and it already uses 51\% fewer fine-mode input tokens than the
hand-tuned baseline at higher quality. Compound execution is the
decisive second tier where topology permits; the first tier carries
deployments where it does not.

\paragraph{The synthesis-only pattern.}
Synthesis groups (no tools, single aggregation agent) pass the quality
floor in standard compound across all models that enter compound
mode. This is the safe, universally applicable deployment
pattern---compound the final aggregation stage only. For mid-tier
models with sufficient quality headroom (Sonnet, GPT-4o-mini),
sequential compound extends viability to all groups.

\paragraph{Quality floor as a deployment parameter.}
We set \texttt{quality\-\_floor=0.75} by calibration against the due
diligence task. This is a design parameter, not a universal
standard. Deployments serving different task types (code generation,
customer support, data extraction) will have different floor
requirements. The framework exposes the floor as a first-class
configuration parameter, and the Pareto sweep provides the empirical
tradeoff curve to inform the choice.

\paragraph{Composition score as a behavioral fingerprint, not a capability ranking.}
The score is not a capability ranking---Gemini-2.5-pro and GPT-4o-mini
score identically (0.177--0.181) despite very different capabilities.
It is a behavioral fingerprint: how does this model use tools and
accumulate coordination overhead in fine-grained mode? Models with
identical fingerprints receive identical switching decisions
regardless of their benchmark performance. A consequence we did not
anticipate: fingerprint stability means the controller's partition
survives model-tier upgrades within a provider's fleet without
retuning, because newer models in the same family tend to inherit the
older model's tool-call structure.

\paragraph{Quality variance across judges as a methodological
constraint.}
Cross-provider judge calibration (Opus min-detectable-difference
0.030; GPT-4o 0.065, see \S\ref{sec:qual_method}) bounds what this
paper's within-provider and cross-provider comparisons can claim. Any
LLM-judged systems result that ignores the judge's noise floor is
reporting measurement noise as effect. We encourage the field to
report min-detectable-difference alongside means.

\paragraph{Future work: knowledge-base compound execution.}
Current two-phase execution injects gather-phase tool results directly
into the reasoning-phase prompt---a flat context dump. A natural
extension replaces this with a shared per-run knowledge store: gather
agents write to the store, reasoning agents query it selectively (a
retrieval layer over tool results). This enables cross-run caching
(warm runs skip the gather phase) and scales to pipelines with 100+
agents, where full context injection becomes prohibitive.

\paragraph{Future work: cross-pipeline competitive benchmarks.}
The competitive benchmarks here cover one pipeline topology per
framework (LangGraph on a 14-agent competitive intelligence pipeline;
DSPy on a 5-agent due diligence pipeline). Repeating each head-to-head
on the other pipeline would test whether the dominance transfers or
whether specific topologies favor specific alternatives.

\section{Conclusion}

We presented Agent Capsules, an adaptive execution framework that
treats multi-agent pipeline execution as an optimization problem with
empirical quality constraints. The runtime instruments coordination
overhead, scores composition opportunity, selects among three compound
execution strategies, and gates every compound decision on
rolling-mean output quality, all without per-model or per-deployment
configuration.

Across five models and three providers, the quality gate preserves
0.26--0.42 quality points that unconstrained batching loses, matching
an LLM-judge-defined oracle on every measured cell without advance
knowledge of model behavior.
Sequential compound with adaptive output guidance delivers 63--64\%
output-token savings on Sonnet and Haiku at quality-neutral cost and
stays neutral on already-terse models, avoiding the 0.160-point
regression that a uniform concise setting produces on flash-class
models. The three-tier escalation ladder is validated by both a
positive result---sequential unlocks every group for Sonnet and
GPT-4o-mini---and a controlled negative result: adding a richer
reasoning pre-pass increases compression pressure rather than
reducing it, confirming that merged calls are the fundamental
bottleneck.

Against a hand-tuned LangGraph implementation of a 14-agent pipeline
and a DSPy implementation of a 5-agent pipeline (evaluated both
uncompiled and after MIPROv2 compilation), Agent Capsules uses
substantially fewer tokens at equal or higher quality: 51\%
fewer fine-mode and 42\% fewer compound-mode input tokens than hand-tuned
LangGraph, at $+$0.020 and $+$0.017 quality; 19\% fewer tokens than
uncompiled DSPy at parity, and 68\% fewer than MIPROv2 at $+$0.052
quality (\S\ref{sec:lg-comparison}). The fine-mode rows of both
benchmarks beat their baselines before compound execution ever
engages: topology-aware context injection, per-group policy
resolution, adaptive output guidance, and cache-aligned prompts
combine to form an orchestration layer that hand-written agent code
does not match without per-pipeline engineering.

The central finding is not that batching saves tokens---that is
arithmetic. It is that a quality gate makes batching trustworthy at
scale, and that the three-tier escalation ladder turns the
quality-savings tradeoff from a binary choice into a continuous one
the runtime navigates on the deployer's behalf.

\bibliographystyle{ACM-Reference-Format}
\bibliography{refs}

@inproceedings{orca,
  title     = {Orca: A Distributed Serving System for Transformer-Based Generative Models},
  author    = {Yu, Gyeong-In and Jeong, Joo Seong and Kim, Geon-Woo and Kim, Soojeong and Chun, Byung-Gon},
  booktitle = {16th USENIX Symposium on Operating Systems Design and Implementation (OSDI)},
  year      = {2022},
}

@inproceedings{vllm,
  title     = {Efficient Memory Management for Large Language Model Serving with PagedAttention},
  author    = {Kwon, Woosuk and Li, Zhuohan and Zhuang, Siyuan and Sheng, Ying and Zheng, Lianmin and Yu, Cody Hao and Gonzalez, Joseph and Zhang, Hao and Stoica, Ion},
  booktitle = {Proceedings of the ACM SIGOPS 29th Symposium on Operating Systems Principles (SOSP)},
  year      = {2023},
}

@inproceedings{flashattention,
  title     = {FlashAttention: Fast and Memory-Efficient Exact Attention with IO-Awareness},
  author    = {Dao, Tri and Fu, Dan and Ermon, Stefano and Rudra, Atri and Ré, Christopher},
  booktitle = {Advances in Neural Information Processing Systems (NeurIPS)},
  year      = {2022},
}

@inproceedings{llmlingua,
  title     = {LLMLingua: Compressing Prompts for Accelerated Inference of Large Language Models},
  author    = {Jiang, Huiqiang and Wu, Qianhui and Lin, Chin-Yew and Yang, Yuqing and Qiu, Lili},
  booktitle = {Proceedings of the 2023 Conference on Empirical Methods in Natural Language Processing (EMNLP)},
  year      = {2023},
}

@article{frugalgpt,
  title   = {FrugalGPT: How to Use Large Language Models While Reducing Cost and Improving Performance},
  author  = {Chen, Lingjiao and Zaharia, Matei and Zou, James},
  journal = {arXiv preprint arXiv:2305.05176},
  year    = {2023},
}

@software{crewai,
  title  = {CrewAI: Framework for Orchestrating Role-Playing Autonomous AI Agents},
  author = {Moura, Joao},
  year   = {2024},
  url    = {https://github.com/joaomdmoura/crewAI},
}

@software{langgraph,
  title  = {LangGraph: Build Resilient Language Agents as Graphs},
  author = {LangChain},
  year   = {2024},
  url    = {https://github.com/langchain-ai/langgraph},
}

@software{googleadk,
  title  = {Google Agent Development Kit},
  author = {Google},
  year   = {2025},
  url    = {https://google.github.io/adk-docs/},
}

@inproceedings{sglang,
  title     = {SGLang: Efficient Execution of Structured Language Model Programs},
  author    = {Zheng, Lianmin and Yin, Liangsheng and Xie, Zhiqiang and Sun, Jeff and Diao, Chuyue and Yu, Theodore and Arfeen, Ishaan and Bhatt, Rohith and Gonzalez, Joseph E. and Stoica, Ion},
  booktitle = {Advances in Neural Information Processing Systems (NeurIPS)},
  year      = {2024},
}

@inproceedings{dspy,
  title     = {DSPy: Compiling Declarative Language Model Calls into State-of-the-Art Pipelines},
  author    = {Khattab, Omar and Singhvi, Arnav and Maheshwari, Paridhi and Zhang, Zhiyuan and Santhanam, Keshav and Vardhamanan, Sri and Haq, Saiful and Sharma, Ashutosh and Joshi, Thomas T. and Moazam, Hanna and Miller, Heather and Zaharia, Matei and Potts, Christopher},
  booktitle = {International Conference on Learning Representations (ICLR)},
  year      = {2024},
}

@inproceedings{routellm,
  title     = {RouteLLM: Learning to Route LLMs with Preference Data},
  author    = {Ong, Isaac and Almahairi, Amjad and Wu, Vincent and Chiang, Wei-Lin and Wu, Tianhao and Gonzalez, Joseph E. and Kadous, M. Waleed and Stoica, Ion},
  booktitle = {International Conference on Learning Representations (ICLR)},
  year      = {2025},
}

@inproceedings{metagpt,
  title     = {MetaGPT: Meta Programming for A Multi-Agent Collaborative Framework},
  author    = {Hong, Sirui and Zhuge, Mingchen and Chen, Jonathan and Zheng, Xiawu and Cheng, Yuheng and Zhang, Ceyao and Wang, Jinlin and Wang, Zili and Yau, Steven Ka Shing and Lin, Zijuan and Zhou, Liyang and Ran, Chenyu and Xiao, Lingfeng and Wu, Chenglin and Schmidhuber, Jürgen},
  booktitle = {International Conference on Learning Representations (ICLR)},
  year      = {2024},
}

@article{agentscope,
  title   = {AgentScope: A Flexible yet Robust Multi-Agent Platform},
  author  = {Gao, Dawei and Ji, Zitao and Chen, Zehao and Tan, Weirui and Li, Shaguo and Xu, Gang and Wang, Zhiyi and Liu, Boyuan and Wang, Xingyuan and Wang, Guorui and others},
  journal = {arXiv preprint arXiv:2402.14034},
  year    = {2024},
}

@inproceedings{gorilla,
  title     = {Gorilla: Large Language Model Connected with Massive APIs},
  author    = {Patil, Shishir G. and Zhang, Tianjun and Wang, Xin and Gonzalez, Joseph E.},
  booktitle = {Advances in Neural Information Processing Systems (NeurIPS)},
  year      = {2024},
}

@inproceedings{toolllm,
  title     = {ToolLLM: Facilitating Large Language Models to Master 16000+ Real-world APIs},
  author    = {Qin, Yujia and Liang, Shengding and Ye, Yining and Zhu, Kunlun and Yan, Lan and Lu, Yaxi and Lin, Yankai and Cong, Xin and Tang, Xiangru and Qian, Bill and Zhao, Sihan and Hong, Lauren and Tian, Runchu and Xie, Ruobing and Zhou, Jie and Gerstein, Mark and Li, Dahai and Liu, Zhiyuan and Sun, Maosong},
  booktitle = {International Conference on Learning Representations (ICLR)},
  year      = {2024},
}

@inproceedings{capsules,
  title     = {Capsules: Expressing Composable Computations in a Parallel Programming Model},
  author    = {Mandviwala, Hasnain A. and Ramachandran, Umakishore and Knobe, Kathleen},
  booktitle = {Proceedings of the 20th International Workshop on Languages and Compilers for Parallel Computing (LCPC)},
  year      = {2007},
}

\end{document}